\documentclass[10pt,twocolumn,letterpaper]{article}

\usepackage[pagenumbers]{wacv} 

\newcommand*{\ShowNotes}{}
\definecolor{darkred}{rgb}{0.7,0.1,0.1}
\definecolor{darkgreen}{rgb}{0.1,0.7,0.1}
\definecolor{dblue}{rgb}{0.2,0.2,0.8}
\definecolor{maroon}{rgb}{0.76,.13,.28}
\definecolor{burntorange}{rgb}{0.81,.33,0}
\definecolor{cyan}{rgb}{0.0,0.7,0.94}
\definecolor{salmon}{rgb}{0.99,0.51,0.46}
\definecolor{green}{rgb}{0.03,0.91,0.43}
\definecolor{forestgreen}{rgb}{0.13,0.55,0.13}
\definecolor{grey}{rgb}{0.4,0.4,0.4}
\definecolor{purple}{rgb}{0.29,0,0.51}
\definecolor{crimson}{rgb}{0.86,0.08,0.24}

\ifdefined\ShowNotes
  \newcommand{\colornote}[3]{{\color{#1}\bf{#2: #3}\normalfont}}
\else
  \newcommand{\colornote}[3]{}
\fi

\usepackage{sg-ibs-macros} 
\usepackage{subcaption}

\usepackage{multirow}
\usepackage{pifont}
\newcommand{\cmark}{\ding{51}}

\usepackage{tikz}

\definecolor{wacvblue}{rgb}{0.21,0.49,0.74}
\usepackage[pagebackref,breaklinks,colorlinks,allcolors=wacvblue]{hyperref}

\def\wacvPaperID{1989} 
\def\confName{WACV}
\def\confYear{2026}

\title{SFMNet: Sparse Focal Modulation for 3D Object Detection}

\author{
    Oren Shrout\\
    Technion, Israel\\
    {\tt\small shrout.oren@campus.technion.ac.il}
    \and
    Ayellet Tal\\
    Technion, Israel\\
    {\tt\small ayellet@ee.technion.ac.il}
    \and
    \tt\small { \href{https://github.com/oshrout/SFMNet}{https://github.com/oshrout/SFMNet}}
}

\begin{document}
\maketitle

\begin{abstract}
We propose SFMNet, a novel 3D sparse detector that combines the efficiency of sparse convolutions with the ability to model long-range dependencies.
While traditional sparse convolution techniques efficiently capture local structures, they struggle with modeling long-range relationships. 
However, these relationships are essential for 3D object detection.
In contrast, transformers are designed to capture these long-range dependencies through attention mechanisms.
But, they come with high computational costs, due to their quadratic query-key-value interactions. 
Furthermore, directly applying attention to non-empty voxels is inefficient due to the sparse nature of 3D scenes.
Our SFMNet is built on a novel Sparse Focal Modulation (SFM) module, which integrates short- and long-range contexts with linear complexity by leveraging a new hierarchical sparse convolution design. 
This approach enables SFMNet to achieve high detection performance with improved efficiency, making it well-suited for large-scale LiDAR scenes. 
We show that our detector achieves state-of-the-art performance on autonomous driving datasets.

\end{abstract}

\section{Introduction}

3D object detection is fundamental for applications in autonomous driving~\cite{nuscenes,chang2019argoverse,kitti1,waymo,once,wilson2023argoverse} and robotics~\cite{thrun2006stanley,zhu2017target}.
Extensive research has explored various methods, including point-based~\cite{qi2017pointnet,qi2017pointnet++,qi2018frustum,pointrcnn,pvrcnn,pvrcnn++,shi2020point,yang20203dssd}, range-based~\cite{fan2021rangedet,sun2021rsn,tian2022fully}, and voxel-based approaches~\cite{chen2019fast,chen2023voxelnext,voxelrcnn,pointpillars,parta2,shrout2023gravos,wang2023dsvt,second,ye2020hvnet,yin2021center,zhang2024safdnet,zhang2024hednet,zheng2021cia,zheng2021se,zhou2018voxelnet,song2023vp}.
Point-based methods process raw point clouds but often suffer from high computational costs or reduced accuracy in large-scale scenes due to point sampling.
Range-based methods project 3D points to apply 2D convolutions, making them computationally efficient. 
However, they often struggle with occlusions, object localization, and size regression errors, as they lack the expressive 3D features necessary for accurate detection.

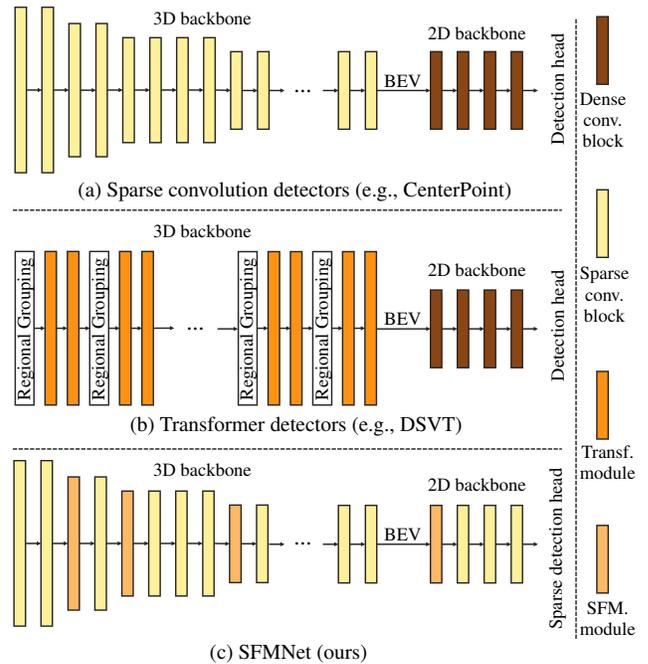
\begin{figure}[t]
    \centering
    \begin{tikzpicture}
        \node[anchor=south west,inner sep=0] at (0,0) {\includegraphics[width=0.95\linewidth]{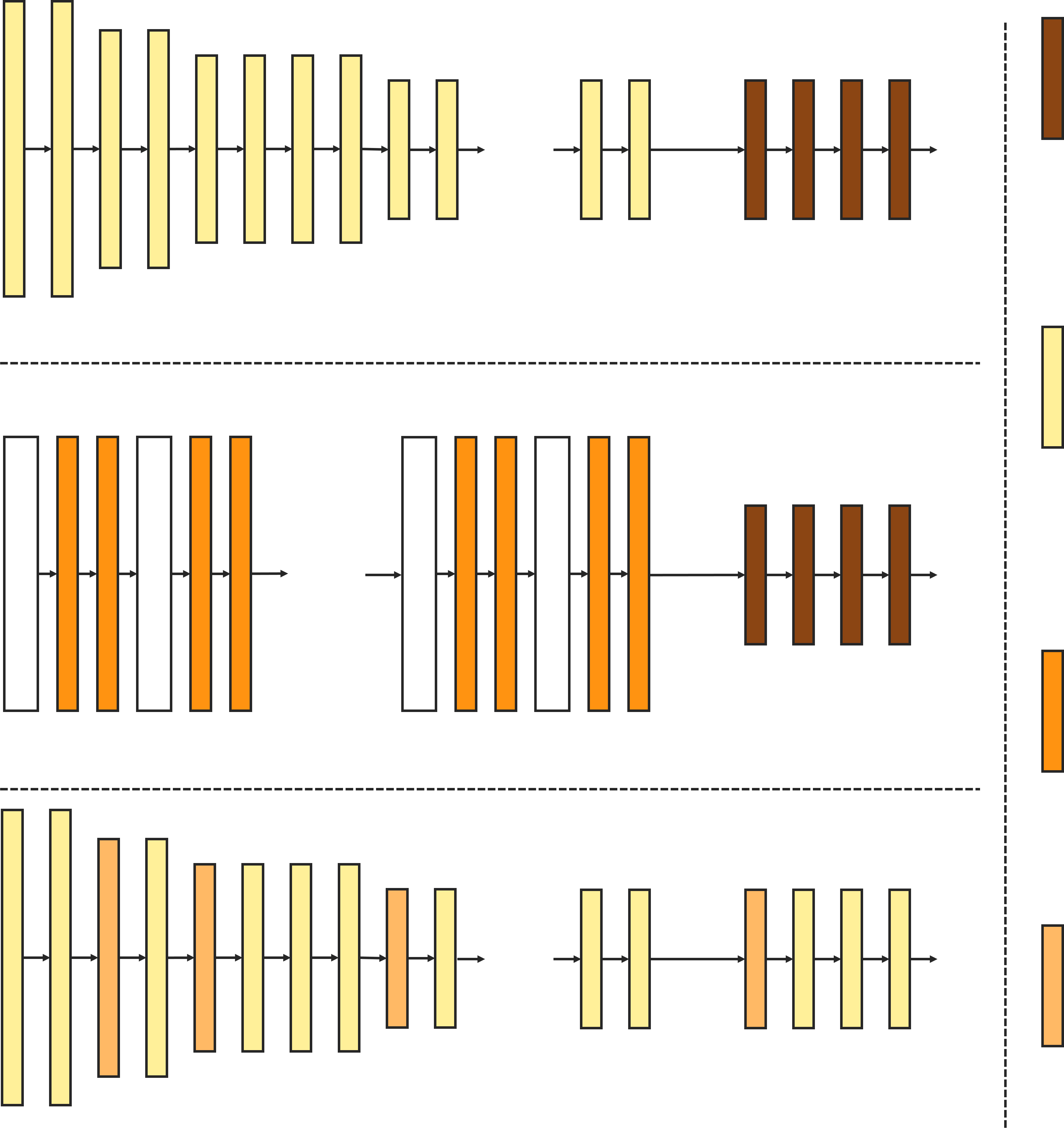}};
        
            
        \node[anchor=west, font=\footnotesize] at (0.75,5.95) {(a) Sparse convolution detectors (e.g., CenterPoint)};
        \node[anchor=west, font=\scriptsize] at (1.75,8.25) {3D backbone};
        \node[anchor=west, font=\scriptsize] at (5.4,8.05) {2D backbone};
        \node[anchor=west, font=\scriptsize] at (4.82,7.41) {BEV};
        \node[anchor=west, font=\scriptsize, rotate=90] at (7.25, 6.45) {Detection head};
        \node[anchor=west, font=\small] at (3.62,7.28) {...};
        
        \node[anchor=west, font=\footnotesize] at (1.44,2.82) {(b) Transformer detectors (e.g., DSVT)};
        \node[anchor=west, font=\scriptsize] at (1.75,5.42) {3D backbone};
        \node[anchor=west, font=\scriptsize] at (5.4,4.9) {2D backbone};
        \node[anchor=west, font=\scriptsize] at (4.82,4.25) {BEV};
        \node[anchor=west, font=\scriptsize, rotate=90] at (7.25, 3.3) {Detection head};
        \node[anchor=west, font=\scriptsize, rotate=90] at (0.18, 3.05) {Regional Grouping};
        \node[anchor=west, font=\scriptsize, rotate=90] at (1.17, 3.05) {Regional Grouping};
        \node[anchor=west, font=\scriptsize, rotate=90] at (3.14, 3.05) {Regional Grouping};
        \node[anchor=west, font=\scriptsize, rotate=90] at (4.14, 3.05) {Regional Grouping};
        \node[anchor=west, font=\small] at (2.19,4.13) {...};

        \node[anchor=west, font=\footnotesize] at (2.5,-0.2) {(c) SFMNet (ours)};
        \node[anchor=west, font=\scriptsize] at (1.75,2.25) {3D backbone};
        \node[anchor=west, font=\scriptsize] at (5.4,2.05) {2D backbone};
        \node[anchor=west, font=\scriptsize] at (4.82,1.39) {BEV};
        \node[anchor=west, font=\scriptsize, rotate=90] at (7.25, 0.0) {Sparse detection head};
        \node[anchor=west, font=\small] at (3.62,1.27) {...};

        \node[anchor=west, font=\scriptsize, align=center] at (7.42,6.9) {Dense\\conv.\\block};
        \node[anchor=west, font=\scriptsize, align=center] at (7.42,4.58) {Sparse\\conv.\\block};
        \node[anchor=west, font=\scriptsize, align=center] at (7.42,2.35) {Transf.\\module};
        \node[anchor=west, font=\scriptsize, align=center] at (7.42,0.27) {SFM.\\module};
        
    \end{tikzpicture}
    \caption{
        {\bf SFMNet in comparison with other architectures.}
        Similar to sparse convolution-based detectors, e.g.,\cite{yin2021center}~(a), our SFMNet detector~(c) comprises a multi-stage 3D backbone with sparse convolution blocks. 
        In contrast, SFMNet is fully sparse, including the 2D backbone, and integrates a novel SFM module (light orange) into both the 3D and 2D backbones.  
        Furthermore, unlike transformers, SFMNet uses sparse convolutions as the token mixer, which makes query modulation efficient.
        Moreover, unlike the single-stride network design used in transformer-based detectors (e.g.,~\cite{wang2023dsvt}), our SFMNet detector employs an encoder-based design.
        }
    \label{fig:teaser}
\end{figure}

Voxel-based methods, which are the focus of this paper, transform irregular point clouds into regular voxel grids. 
This enables the use of convolutions, specifically sparse convolutions, that operate efficiently despite the presence of many empty voxels~\cite{choy20194d,sparseconv,second}. 
While effective at capturing local structures, methods based on submanifold sparse convolution~\cite{graham2017submanifold} substantially limit the receptive field, thereby reducing the model's ability to capture long-range dependencies, which are critical for detection~\cite{wang2018non}. 
In 3D detection, long-range dependencies encompass relationships among voxels within an object as well as between the object and its immediate surroundings.
Given the inherent sparsity and frequent occlusions in LiDAR data, relying only on local features makes detecting small objects, such as pedestrians, extremely difficult when only a few points are available. 
Contextual information, such as the presence of a crosswalk, can greatly aid detection, as even a few points above it may indicate a pedestrian's presence, highlighting the importance of long-range dependencies.

To address these limitations, the use of transformers has been proposed~\cite{chen2023focalformer3d,fan2022embracing,sun2022swformer,vaswani2017attention,wang2023dsvt,wang2023unitr,zhou2022centerformer}.
With self-attention (SA) as their core mechanism, transformers can capture global relationships, increasing the detector representation capacity. 
Despite its benefits, the quadratic complexity of SA is a significant bottleneck, particularly in autonomous vehicle scenarios where LiDAR scenes span hundreds of meters and contain hundreds of thousands of points.

We propose a 'compromise' between the two approaches by using large kernels in convolutional neural networks. This strategy has demonstrated efficiency, primarily in 2D~\cite{ding2022scaling,ding2024unireplknet,liu2022more}.
Large-kernel methods improve representation capacity by increasing the receptive field and offering greater robustness compared to transformers~\cite{chen2024revealing}.
However, this approach is often impractical for 3D scenes, as the number of parameters grows cubically, necessitating techniques like sophisticated kernel weight partitioning with shared weights~\cite{chen2023largekernel3d}, data-driven kernel generation~\cite{lu2023link}, or weight pruning~\cite{feng2024lsk3dnet}.
Although these methods reduce parameter bottlenecks and enlarge the effective receptive field, they offer minimal accuracy improvements and are less efficient than traditional sparse convolutions.

Inspired by these solutions, this paper aims to combine the best of both worlds---enabling transformer-like long-range dependencies while maintaining the efficiency of sparse convolutions.
We introduce a new 3D sparse detector, termed {\em SFMNet}, that relies entirely on (sparse) convolutions; however, between these convolutions, we introduce a novel 'transformer-like' module where the attention mechanism is also replaced with convolutions.
Our {\em SFM} module captures both short- and long-range relationships between voxels, bridging gaps left by traditional sparse convolution detectors while providing greater efficiency than transformer-based detectors.
As illustrated in \figref{fig:teaser}, our model differs from traditional sparse convolution detectors by being fully sparse, including its 2D backbone, and by incorporating our SFM module within the backbones.

The {\em Sparse Focal Modulation (SFM)} module modulates short- and long-range contexts and integrates them with query tokens, similar to transformers. 
However, instead of the quadratic complexity of self-attention, SFM operates with linear complexity relative to the number of tokens (voxels). 
The core idea is to efficiently extract local contexts using sparse convolutions in a hierarchical structure, enabling the design to capture global relationships.
Limiting attention to local context windows has been shown to reduce the complexity bottleneck~\cite{liu2021swin}. 
However, due to the sparse nature of LiDAR points, the number of occupied voxels in each local window can vary significantly, making efficient local-based attention non-trivial~\cite{wang2023dsvt}.
To efficiently extract context windows, we design our SFM module with stacked submanifold sparse convolutions, using small kernels and varying dilation rates. 
This design enables hierarchical context aggregation, progressively expanding the receptive field at each level.
Sparse convolutions capture local structures efficiently, while the hierarchical design and increased dilations allow for learning sparse long-range dependencies. 

We validate the effectiveness of our detector on three well-known autonomous driving datasets.
Our results demonstrate that the network advances the state-of-the-art in long-range detection (Argoverse2~\cite{wilson2023argoverse}), 
for which it is designed. 
On mid- to short-range datasets, Waymo Open~\cite{waymo} and nuScenes~\cite{nuscenes}, it performs comparably or slightly better than previous state-of-the-art detectors.

Thus, the main contributions are twofold:
\begin{itemize}
    \item 
    We propose a novel sparse focal modulation (SFM) module that enables long-range dependencies like transformers while maintaining the efficiency of sparse convolutions. It hierarchically aggregates contextual information from sparse inputs, outperforming traditional attention methods by reducing computational overhead and efficiently handling sparse voxels of varying densities.
    
    
    \item 
    We introduce SFMNet, a 3D object detector built on top of the SFM module, which effectively handles large point cloud scenes. 
    We validate the detector's usefulness on well-known large-scale LiDAR datasets, achieving state-of-the-art results.

\end{itemize}

\section{Related work}

\textbf{3D detection on point clouds.}
3D detection methods are broadly categorized into point-based methods and voxel-based methods. 
Point-based methods use PointNet~\cite{qi2017pointnet,qi2017pointnet++}-like architectures to extract geometric features directly from raw point clouds~\cite{qi2017pointnet,qi2017pointnet++,qi2018frustum,pointrcnn,pvrcnn,pvrcnn++,shi2020point,yang20203dssd,wang2023sat}.
However, in large-scale point cloud scenes typical of outdoor LiDAR datasets in autonomous driving, these methods suffer from lower accuracy and high processing times due to point sampling and neighbor searching.
In contrast, voxel-based methods convert irregular point clouds into regular voxel grids and apply sparse convolutions on non-empty voxels~\cite{zhou2018voxelnet,chen2019fast,voxelrcnn,pointpillars,parta2,second,ye2020hvnet,yin2021center,zheng2021cia,zheng2021se,chen2023voxelnext,zhang2024hednet,zhang2024safdnet,shrout2023gravos,liu2024sparsedet}. 
While efficient for processing large-scale data, these methods rely on submanifold sparse convolution, which might limit network representation capacity by prioritizing efficiency over capturing long-range dependencies.

\noindent
\textbf{Long-range dependencies for 3D detection.}
Recent 3D detection methods capture long-range dependencies using large-kernel sparse convolutions or transformers.
Large-kernel convolution approaches expand the receptive field by increasing kernel sizes, a technique that has shown accuracy gains in 2D detection~\cite{ding2024unireplknet,ding2022scaling,liu2022more}.
However, directly applying large kernels to 3D scenes incurs a cubically increasing computational overhead.
To address this, LargeKernel3D~\cite{chen2023largekernel3d} introduces a spatial partitioning convolution with shared weights among neighboring voxels.
LinK~\cite{lu2023link} further optimizes efficiency by generating sparse kernels dynamically, assigning weights only to non-empty voxels rather than maintaining a dense kernel matrix. 
Similarly, LSK3DNet~\cite{feng2024lsk3dnet} reduces computation by pruning redundant weights across spatial and channel dimensions. 
While these methods alleviate some computational demands, they still face performance challenges in large-scale 3D data.

Transformer-based detectors, by contrast, use self-attention to capture long-range dependencies, enabling global token interactions.
However, applying attention in high-resolution 3D scenes with many tokens is computationally prohibitive~\cite{zhou2022centerformer,chen2023focalformer3d,wang2023unitr}.
To mitigate this, VoTr~\cite{mao2021voxel} implements local and dilated attention, focusing on sparse voxel interactions. 
SWFormer~\cite{sun2022swformer} utilizes a window-based transformer with a bucketing scheme to manage windows of varying non-empty voxels.
SST~\cite{fan2022embracing} refines token processing by grouping tokens regionally, applying attention within each group, and shifting regions to capture structured interactions.
DSVT~\cite{wang2023dsvt} sequentially applies windowed self-attention along the X and Y axes to capture dependencies with reduced complexity. 
Although they improve long-range relationships, they still face quadratic scaling issues with context-window tokens due to self-attention.

We propose a network that combines the strengths of sparse convolutions and transformers. 
Our design maintains the efficiency of sparse convolutions while enabling long-range dependencies similar to transformers.
However, unlike traditional transformers, our module is built from sparse convolutions and operates with linear complexity relative to the number of tokens (voxels).
Furthermore, eliminating the need for sampling helps us achieve better accuracy.

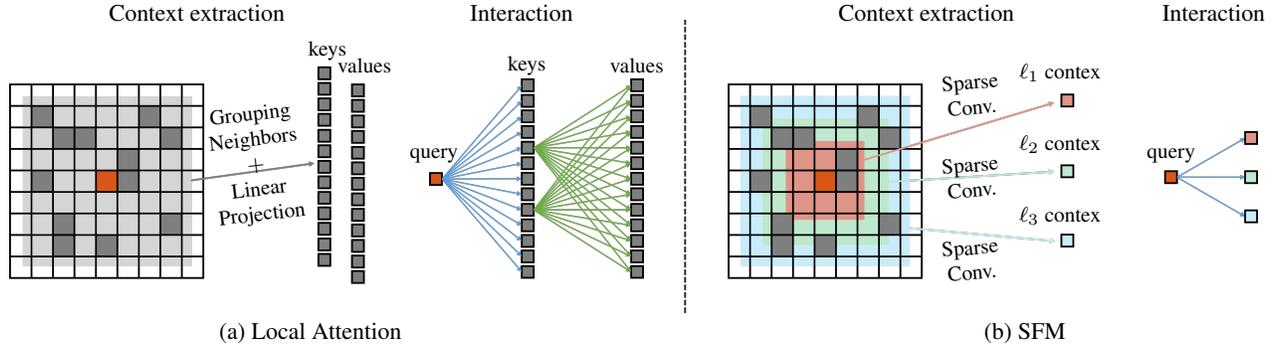
\begin{figure*}[t]
    \centering
    \begin{tikzpicture}
        \node[anchor=south west,inner sep=0] at (0,0) 
        {\includegraphics[width=0.95\linewidth]{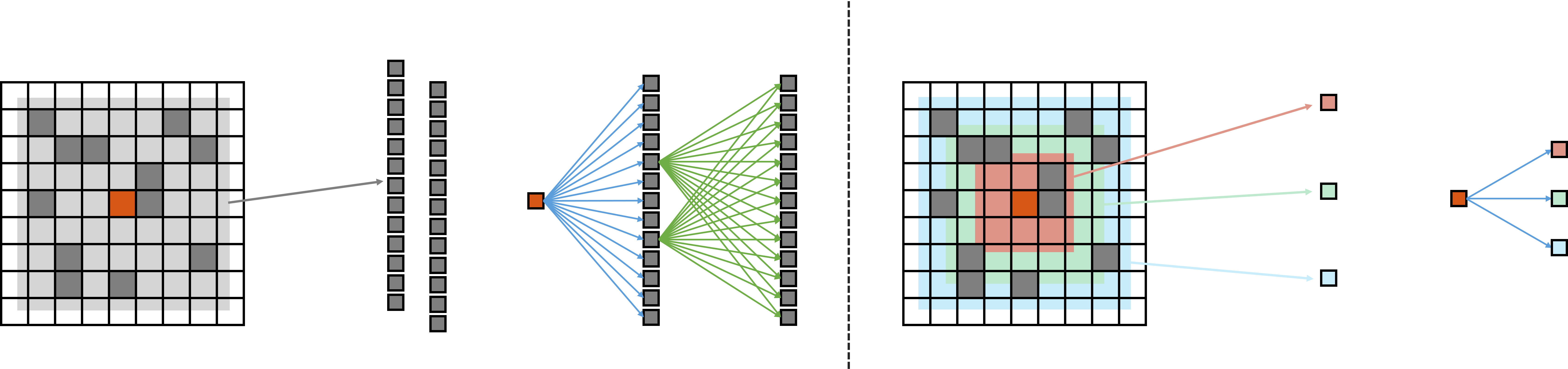}};
        
            
        \node[anchor=north, font=\small] at (4.0,0.0) {(a) Local Attention};
        \node[anchor=west, font=\small] at (1.2,4.0) {Context extraction};
        \node[anchor=west, font=\small] at (6.0,4.0) {Interaction};
        \node[anchor=west, font=\footnotesize] at (3.85,3.5) {keys};
        \node[anchor=west, font=\footnotesize] at (4.24,3.3) {values};
        \node[anchor=west, font=\footnotesize] at (5.2,2.1) {query};
        \node[anchor=west, font=\footnotesize] at (6.5,3.3) {keys};
        \node[anchor=west, font=\footnotesize] at (7.86,3.3) {values};
        \node[anchor=north, font=\footnotesize, align=center, rotate=7] at (3.17,2.88) {Grouping\\Neighbors\\$+$\\Linear\\Projection};

        \node[anchor=north, font=\small] at (13.5,0.0) {(b) SFM};
        \node[anchor=west, font=\small] at (10.9,4.0) {Context extraction};
        \node[anchor=west, font=\small] at (15.2,4.0) {Interaction};
        \node[anchor=west, font=\footnotesize] at (13.3,3.2) {$\ell_1$ contex};
        \node[anchor=west, font=\footnotesize] at (13.3,2.25) {$\ell_2$ contex};
        \node[anchor=west, font=\footnotesize] at (13.3,1.3) {$\ell_3$ contex};
        \node[anchor=west, font=\footnotesize] at (15.0,2.1) {query};
        \node[anchor=north, font=\footnotesize, align=center, rotate=5] at (12.75,3.3) {Sparse\\Conv.};
        \node[anchor=north, font=\footnotesize, align=center, rotate=5] at (12.75,2.2) {Sparse\\Conv.};
        \node[anchor=north, font=\footnotesize, align=center, rotate=5] at (12.75,1.1) {Sparse\\Conv.};
        
    \end{tikzpicture}
    \caption{
        {\bf SFM module vs. Local attention.}
        Given a query token (burnt orange), local-based attention~(a) identifies and groups non-empty tokens (voxels) within a context window (light gray) to form key and value tokens.
        The query-key-value interactions have quadratic complexity. 
        In contrast, SFM~(b) leverages sparse convolutions to extract multiple levels of context, reducing query interactions to only a few focal contexts and achieving linear complexity with respect to the number of non-empty voxels in the context window.
        Furthermore, while local-based attention suffers from inefficiencies in batch processing due to the varying numbers of non-empty voxels within each context window, our SFM approach is robust to this variability, extracting a constant number of contexts for each window.
        }
	\label{fig:transformer_vs_sfm}
\end{figure*}

\section{Method}

Voxel-based detectors are widely used for 3D detection in outdoor scenes, such as those in autonomous driving scenarios.
They commonly leverage sparse convolutions for efficient learning in sparse environments. 
There are two main types of sparse convolutions: regular~\cite{graham2015sparse} and submanifold~\cite{graham2017submanifold}. 
Regular sparse convolutions expand features into (not necessarily occupied) neighboring voxels, enabling information exchange between spatially disconnected regions, but increasing feature map density. 
Their expansive nature makes them suitable for downsampling and upsampling.
Submanifold sparse convolutions operate only on occupied voxels, ensuring that the output retains the same sparsity as the input, thereby facilitating efficient learning of local structures. 
However, due to their limited kernel size, they lack long-range dependencies, which are useful for object detection~\cite{wang2018non}.

We aim to design an efficient detector built from sparse convolutions capable of capturing both short- and long-range dependencies.
While transformers are effective for modeling global relationships in 2D, applying them to sparse point clouds is challenging: 
(1) With numerous occupied voxels in each scene, standard self-attention suffers from massive query-key and query-value interactions, leading to a high computational overhead or suboptimal accuracy due to subsampling~\cite{pan20213d,zhao2021point}. 
(2) The varying voxel density across regions hinders efficient batching~\cite{fan2022embracing,wang2023dsvt}.

To address both the computational overhead and the varying number of occupied voxels within a context window, we propose replacing self-attention with {\em Sparse Focal Modulation (SFM)} as the token mixer in the transformer.
The SFM module reduces query interactions to only a few focal contexts around each query by hierarchically extracting representations of the surrounding connections. 
Additionally, sparse convolutions were designed to efficiently handle varying numbers of occupied voxels within each local region and can facilitate effective context extraction.

Our SFM module is described in~\secref{subsection:sfm}.
\figref{fig:transformer_vs_sfm} illustrates its context extraction and interaction phases compared to naive local attention.
Local attention involves extracting a context for each query token, then performing query-key and query-value interactions.
This requires identifying all non-empty voxels in the context window, sampling or padding them to a fixed maximum, and concatenating features for batch processing.
In high-resolution scenes with hundreds of thousands of voxels, this quickly becomes a bottleneck, and query-key-value interactions incur quadratic complexity with the number of tokens per window.
In contrast, SFM modulates query interactions by extracting multiple context levels via sparse convolutions, capturing short-range structures at lower levels and long-range dependencies at higher levels.
These sparse convolutions efficiently handle varying numbers of non-empty voxels within context windows, enabling effective batch processing.
By using convolutions with different receptive fields for context extraction, the interaction phase is reduced to only the number of context levels.

The effectiveness of our proposed SFM is demonstrated through a novel sparse detector, called {\em SFMNet,} introduced in~\secref{subsection:sfmnet}. 
Our detector effectively captures long-range dependencies with a large receptive field, improving performance in object detection tasks.

\subsection{Sparse focal modulation}
\label{subsection:sfm}

Focal Modulation (FM) is an alternative token mixer to self-attention, designed to limit token interactions. 
It has been applied in 2D tasks like classification, segmentation, and detection~\cite{yang2022focal,ma2024efficient}. 
Extending FM to 3D faces two primary challenges.
First, 3D representations are inherently sparse, with voxel count scaling cubically. 
For instance, a typical Argoverse2~\cite{wilson2023argoverse} scene spans $400 \times 400 \times 8$ meters, yielding approximately $640$ million voxels at $0.1 \times 0.1 \times 0.2$ m resolution. 
With only about 100,000 sampled points, this results in a high percentage of empty voxels.
Second, 2D FM typically captures global context using stacked depth-wise convolutions with progressively larger kernels to model multi-scale visual features. 
However, in 3D detection, sparse depth-wise convolutions offer no benefit over spatial ones and are memory-intensive~\cite{chen2023largekernel3d}.
Furthermore, applying large kernels in 3D scenes leads to a cubic increase in parameter count, substantially raising computational costs.

We propose a new {\em Sparse Focal Modulation (SFM)} module that addresses these challenges and captures both short- and long-range contexts while adaptively exchanging information with each token query. 
Unlike the conventional approach of using large kernels to expand the context window densely, we expand the context window sparsely with small kernels and increased dilation rates. 
We reason that queries related to small objects often rely on local structures and benefit from compact, focused context windows.
In contrast, queries for large objects (e.g., partially occluded vehicles or roads) require broader but not necessarily dense context windows, given the sparsity of LiDAR point clouds. 
Hence, our context aggregation is sparse.

\begin{figure}[t]
    \centering
    \begin{tikzpicture}
        \node[anchor=south west,inner sep=0] at (0,0) 
        {\includegraphics[width=0.90\linewidth]{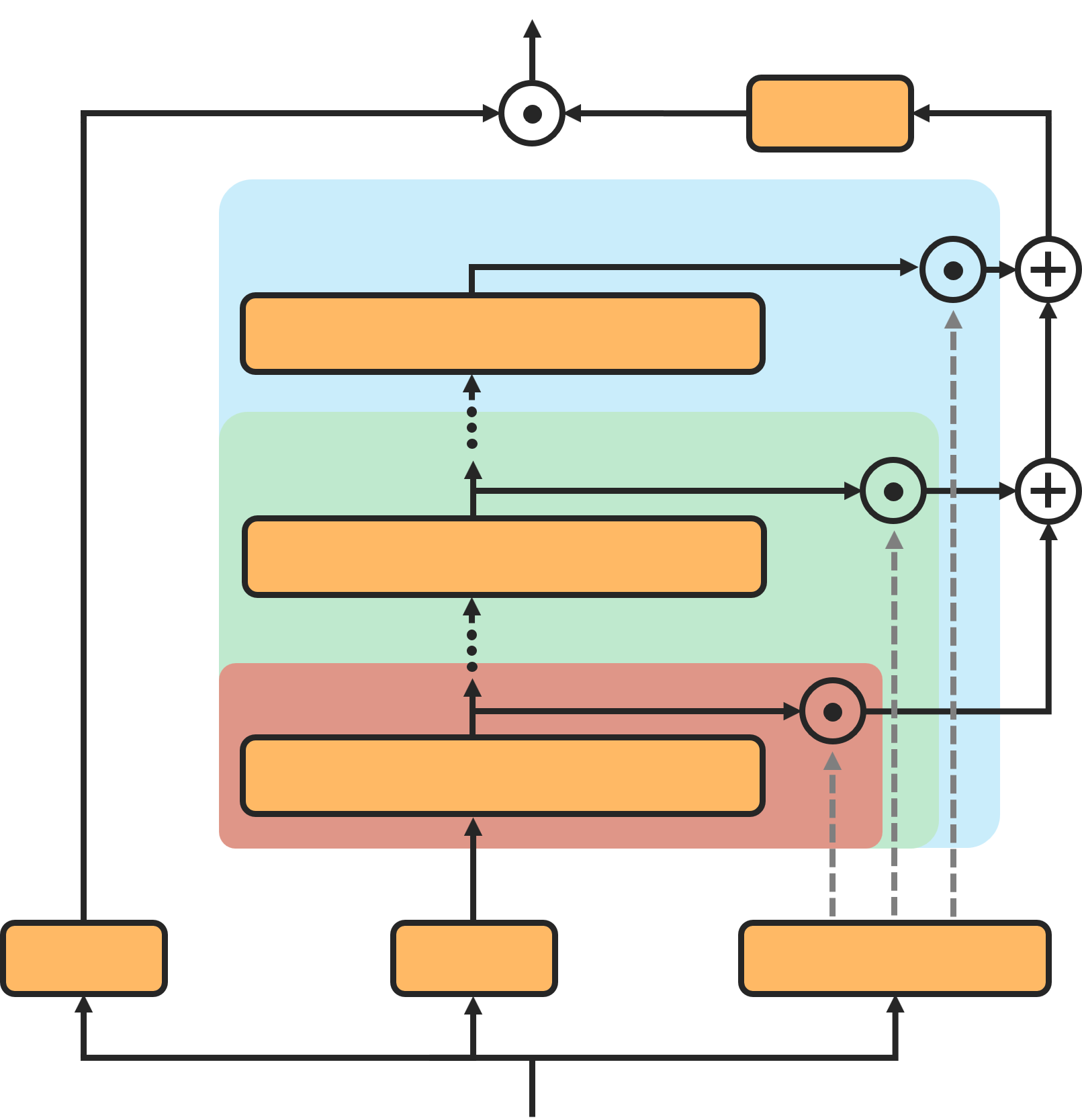}};
        
            
        \node[anchor=west, font=\small] at (1.6,6.35) {\textbf{Context Aggregation}};
        \node[anchor=west, font=\small] at (2.1,2.36) {$\text{SubMConv}(\cdot;k_1,d_1)$};
        \node[anchor=west, font=\footnotesize] at (2.3,2.89) {GeLU};
        \node[anchor=west, font=\small] at (2.1,3.89) {$\text{SubMConv}(\cdot;k_\ell,d_\ell)$};
        \node[anchor=west, font=\footnotesize] at (2.3,4.4) {GeLU};
        \node[anchor=west, font=\small] at (2.02,5.43) {$\text{SubMConv}(\cdot;k_L,d_L)$};
        \node[anchor=west, font=\footnotesize] at (2.3,5.95) {GeLU};

        \node[anchor=west, font=\small] at (0.25,1.13) {FC};
        \node[anchor=west, font=\small] at (3.0,1.13) {FC};
        \node[anchor=west, font=\small] at (5.8,1.13) {Gate};
        \node[anchor=west, font=\small] at (5.4,6.98) {$h(\cdot)$};

        \node[anchor=west, font=\small] at (3.25,0.25) {$x$};
        \node[anchor=west, font=\small] at (0.15,1.65) {$q$};
        \node[anchor=west, font=\small] at (2.7,1.65) {$f^0$};
        \node[anchor=west, font=\small] at (4.35,4.6) {$f^\ell$};
        \node[anchor=west, font=\small] at (5.65,3.7) {$g^\ell$};
        \node[anchor=west, font=\small] at (4.21,7.2) {ctx};
        \node[anchor=west, font=\small] at (3.26,7.33) {$z$};
        
    \end{tikzpicture}
    \caption{
        {\bf Sparse Focal Modulation (SFM).}
        SFM leverages sparse convolutions to aggregate multi-level contexts and modulate them with query features, efficiently capturing both local and global dependencies.
        }
	\label{fig:FocalModulation}
\end{figure}

\figref{fig:FocalModulation} illustrates our SFM module, which modulates each non-empty voxel feature ($x$) with surrounding information to produce a spatially aware feature ($z$).
Its core component, the {\em Context Aggregation (CA)} block, captures context for each query token by:
1) efficiently extracting contexts hierarchically, from local to global regions (e.g., red, green, and blue regions in~\figref{fig:transformer_vs_sfm});
and 
2) adaptively aggregating context features at each level using a gating mechanism to provide meaningful context for each query.
The aggregated context of the CA block, ctx, is modulated with the corresponding query $q$ to generate the feature $z$.

Submanifold sparse convolutions efficiently handle voxel sparsity, as they operate strictly on non-empty voxels.
By stacking submanifold sparse convolutions with small kernels, we can effectively capture local contexts at each level. 
Specifically, given a feature vector $x \in \mathbb{R}^{N \times C}$, where $N$ represents the number of voxels and $C$ denotes the number of features, we project this vector into new feature spaces using linear layers to generate both per-voxel queries $q \in \mathbb{R}^{N \times C}$ and initial focal features $f^0 \in \mathbb{R}^{N \times C}$. 
The focal features are then fed into the Context Aggregation (CA) module.
The CA produces $L$ levels of focal feature maps, $f^\ell$, with each level fed to a submanifold sparse filter that extracts the next focal context. 
Each level has a progressively larger effective receptive field than the previous level.
At level $\ell$, the focal feature $f^\ell \in \mathbb{R}^{N \times C}$ is computed as
\begin{equation}
    f^\ell = \text{GeLU}(\text{SubMConv}(f^{\ell-1};k^\ell,d^\ell)),
\end{equation}
where $\text{SubMConv}(\cdot;k^\ell,d^\ell)$ denotes a submanifold sparse convolution with kernel size $k^\ell$ and dilation rate $d^\ell$, and GeLU is an activation function~\cite{hendrycks2016gaussian}. 
The hierarchical convolutions effectively capture multiple levels of focal abstraction, with the effective receptive field at level $\ell$ given by: 
\begin{equation}
    r^\ell = 1 + \sum_{i=1}^{\ell} (k^i - 1) \cdot d^i.
\end{equation}

When aggregating context features at each level, we note that not all queries benefit equally from short-range or long-range contexts. 
Specifically, queries for small objects are more likely to gain from the short-range context available at lower focal levels, while larger objects or background regions tend to benefit more from the long-range context captured at higher focal levels.
To address this, a gating mechanism is employed to weigh the focal features from the different levels.
This gate adaptively regulates the amount of context extracted from each level, enabling the CA module to learn suitable contexts for various queries. 
By combining the focal features with the gate weights, the CA produces a context feature vector for each query, defined as:
\begin{equation}
    \text{ctx} = \text{h}(\sum_{\ell=1}^L f^\ell \cdot g^\ell) \in \mathbb{R}^{N \times C},
\end{equation}
where $g^\ell \in \mathbb{R}^{N \times 1}$ represents the gate weight for focal level $\ell$, and $h(\cdot)$ is a linear layer that projects the output context into the query space. 

Finally, we modulate the focal contexts (ctx) to the queries ($q$) through element-wise multiplication, which enables each query to learn its relevant context:
\begin{equation}
    z = q \cdot \text{ctx} \in \mathbb{R}^{N \times C}.
\end{equation}


In summary, to address voxel sparsity in 3D space, we propose Sparse Focal Modulation (SFM), which expands the context window efficiently using submanifold sparse convolutions with small kernels and increasing dilation. 
To handle varying contextual needs, the Context Aggregation (CA) block employs a gating mechanism to adaptively aggregate multi-scale features, allowing each query to focus on the most relevant context for accurate 3D detection.


To demonstrate its utility, we incorporated a single SFM module into each backbone (3D and 2D) within the CenterPoint~\cite{yin2021center} architecture, without otherwise altering the structure. 
This has already yielded a $2.6\%$ improvement in Level-2 mAPH on a $20\%$ subset of the Waymo Open Dataset~\cite{waymo}.

\subsection{SFMNet}
\label{subsection:sfmnet}

Building on our sparse focal modulation (SFM) module, we present SFMNet, a simple and efficient sparse 3D object detector. 
It enables learning long-range dependencies that convolution-based backbones lack, while maintaining sparse convolution efficiency.


Our architecture is built on two core principles. 
First, increasing the effective receptive field (ERF) in mid- and high-level layers enhances the network’s ability to capture broader spatial context—this is primarily achieved using a single SFM block. 
Second, once a sufficient receptive field is established, adding depth with SRB blocks becomes more efficient, as they act like small kernels and offer better parameter utilization. 
In practice, we find that one SFM block combined with one SRB block is sufficient to meet our design goals. 
When further deepening the network, we prioritize adding more SRB blocks rather than additional SFM blocks (see \secref{sec:ablation}). 
This design choice aligns with the observation from \cite{ding2024unireplknet} that, after the receptive field is adequately covered, small kernels become more efficient than large ones—an analogy that maps well to our use of SFM (large kernel equivalent) and SRB (small kernel equivalent) blocks.

\begin{figure}[t]
    \centering
    \begin{tikzpicture}
        \node[anchor=south west,inner sep=0] at (0,0) 
        {\includegraphics[width=0.95\linewidth]{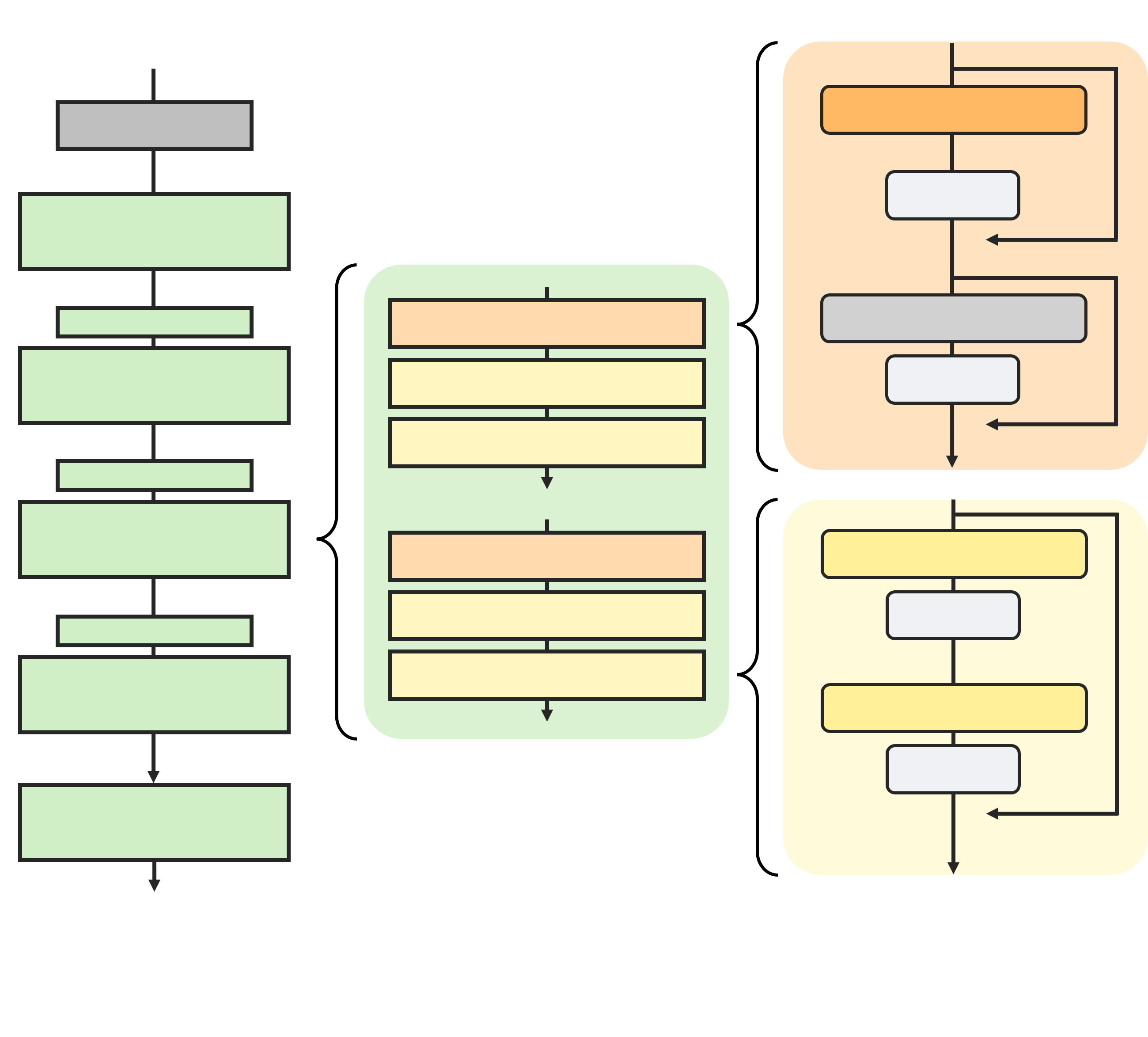}};
        
            
        \node[anchor=west, font=\small] at (0.25,6.9) {Point cloud};
        \node[anchor=west, font=\scriptsize] at (0.68,6.28) {VFE};
        \node[anchor=west, font=\scriptsize] at (0.58,5.54) {Stage 1};
        \node[anchor=west, font=\tiny] at (0.78,4.92) {$2\times$};
        \node[anchor=west, font=\scriptsize] at (0.58,4.48) {Stage 2};
        \node[anchor=west, font=\tiny] at (0.78,3.87) {$2\times$};
        \node[anchor=west, font=\scriptsize] at (0.58,3.42) {Stage 3};
        \node[anchor=west, font=\tiny] at (0.78,2.8) {$2\times$};
        \node[anchor=west, font=\scriptsize] at (0.58,2.34) {Stage 4};
        \node[anchor=west, font=\tiny] at (0.31,1.94) {To BEV};
        \node[anchor=west, font=\scriptsize] at (0.31,1.5) {2D backbone};
        \node[anchor=west, font=\scriptsize] at (0.2,0.86) {Detection head};

        \node[anchor=west, font=\scriptsize] at (3.1,4.92) {SFM Block};
        \node[anchor=west, font=\scriptsize] at (3.45,4.5) {SRB};
        \node[anchor=west, font=\scriptsize] at (3.45,4.1) {SRB};
        \node[anchor=west, font=\scriptsize] at (3.56,3.7) {...};
        \node[anchor=west, font=\scriptsize] at (3.1,3.32) {SFM Block};
        \node[anchor=west, font=\scriptsize] at (3.45,2.9) {SRB};
        \node[anchor=west, font=\scriptsize] at (3.45,2.5) {SRB};

        \node[anchor=west, font=\tiny] at (6.24,6.7) {$x$};
        \node[anchor=west, font=\scriptsize] at (6.25,6.4) {SFM};
        \node[anchor=west, font=\tiny] at (6.24,6.1) {$z$};
        \node[anchor=west, font=\scriptsize] at (6.3,5.8) {LN};
        \node[anchor=west, font=\scriptsize] at (6.5,5.5) {+};
        \node[anchor=west, font=\tiny] at (6.22,5.28) {$y'$};
        \node[anchor=west, font=\scriptsize] at (6.19,4.95) {MLP};
        \node[anchor=west, font=\scriptsize] at (6.3,4.53) {LN};
        \node[anchor=west, font=\scriptsize] at (6.5,4.24) {+};
        \node[anchor=west, font=\tiny] at (6.24,4.1) {$y$};
        
        \node[anchor=west, font=\scriptsize] at (5.94,3.34) {SubMConv};
        \node[anchor=west, font=\scriptsize] at (6.3,2.91) {BN};
        \node[anchor=west, font=\tiny] at (6.0,2.6) {ReLU};
        \node[anchor=west, font=\scriptsize] at (5.94,2.28) {SubMConv};
        \node[anchor=west, font=\scriptsize] at (6.3,1.85) {BN};
        \node[anchor=west, font=\scriptsize] at (6.5,1.55) {+};
        \node[anchor=west, font=\tiny] at (6.0,1.34) {ReLU};

        \node[anchor=west, font=\scriptsize] at (0.3,0.1) {(a) SFMNet};
        \node[anchor=west, font=\scriptsize] at (2.7,0.1) {(b) Stage structure};
        \node[anchor=west, font=\scriptsize] at (5.35,0.1) {(c) SFM Block \& SRB};
    \end{tikzpicture}
    \caption{
        {\bf SFMNet architecture.}
        (a)~The  SFMNet architecture extracts voxel features from a point cloud input (VFE), processes them through multiple stages, and converts them to a bird's-eye view (BEV) representation before passing the output to a 2D backbone and detection head.
        (b)~Each stage consists of multiple Sparse Focal Modulation (SFM) blocks followed by Sparse Residual Blocks (SRBs).
        (c)~The SFM block aggregates contextual information and processes it,
        while the SRB employs SubMConv layers with batch norm (BN) and ReLU activations to capture spatial features.
        }
        
	\label{fig:SFMNet}
\end{figure}
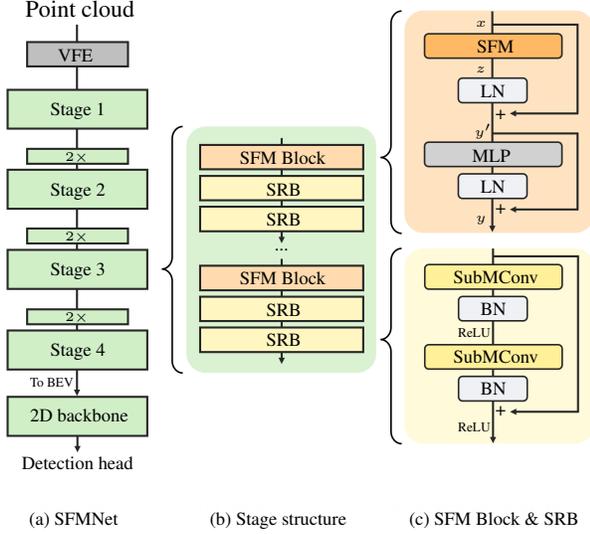

\figref{fig:SFMNet}(a) illustrates the SFMNet architecture.
The input point cloud is voxelized, and features are extracted using a dynamic voxel feature extraction (VFE) method~\cite{zhou2020end}. 
These features are then processed by our 3D and 2D sparse backbones.
Our 3D backbone includes four stages and between each pair of consecutive stages, a regular sparse convolution with stride $2$ is used for downsampling.
Each stage includes an SFM block followed by multiple submanifold residual blocks (SRBs) for high-level feature extraction (\figref{fig:SFMNet}(b)). 
We use a variable number of SFM blocks and SRBs to enhance representational capacity and increase depth depending on the dataset. 
The higher the stage, the more SRB blocks per SFM block should be added to scale the model.

The SFM block employs our SFM module as the token mixer within a standard MetaFormer-like design~\cite{yu2022metaformer}, as illustrated in \figref{fig:SFMNet}(c), where non-empty voxels serve as the tokens.
Formally, given an input feature map $x \in \mathbb{R}^{N \times C}$ and its corresponding focal-modulated feature $z \in \mathbb{R}^{N \times C}$ generated by the SFM module, as described in~\secref{subsection:sfm}, the SFM block is formulated as: 
\begin{align}
    y' = \text{LN}(z) + x, \quad
    y = \text{LN}(\text{MLP}(y')) + y',
\end{align} 
where LN denotes LayerNorm, the multilayer perceptron (MLP) has a single hidden layer, and $y$ is the refined feature from focal sparse modulation. The 3D sparse features are then compressed into 2D bird's-eye view (BEV) features, as in~\cite{zhang2024safdnet}, and passed to a 2D backbone. Like the 3D backbone, the 2D backbone includes an SFM block followed by SRBs, but uses 2D sparse convolutions.

\section{Experiments}

\textbf{Datasets and metrics.}
To validate our approach's effectiveness, we conducted experiments on three popular datasets: Argoverse2~\cite{wilson2023argoverse}, Waymo Open~\cite{waymo}, and nuScenes~\cite{nuscenes}.
The detection range of these datasets varies, with long-range detection in Argoverse2 (up to $\pm 200$ meters), mid-range in Waymo Open (up to $\pm 75$ meters), and short-range in nuScenes (up to $\pm 54$ meters).
We use the standard evaluation metrics for each dataset:
mAP for Argoverse2, mAP and mAP weighted by heading accuracy (mAPH) for Waymo Open, and mAP and nuScenes detection score (NDS) for nuScenes.
On Waymo, the metrics are applied to two difficulty levels: L1 for objects with more than five LiDAR points and L2 for those with at least one sampled point.

\begin{table*}[t]
    \scriptsize
    \setlength\tabcolsep{2pt} 
    \centering
    \begin{tabular}{l | c | c c c c c c c c c c c c c c c c c c c c c c c c c c}
         \toprule
            Method & \rotatebox{90}{mAP} & \rotatebox{90}{Vehicle} & \rotatebox{90}{Bus} & \rotatebox{90}{Pedestrian} & \rotatebox{90}{Stop Sign} & \rotatebox{90}{Box Truck} & \rotatebox{90}{Bollard} & \rotatebox{90}{C-Barrel} & \rotatebox{90}{Motorcyclist} & \rotatebox{90}{MPC-Sign} & \rotatebox{90}{Motorcycle} & \rotatebox{90}{Bicycle} & \rotatebox{90}{A-Bus} & \rotatebox{90}{School Bus} & \rotatebox{90}{Truck Cab} & \rotatebox{90}{C-Cone} & \rotatebox{90}{V-Trailer} & \rotatebox{90}{Sign} & \rotatebox{90}{Large Vehicle} & \rotatebox{90}{Stroller} & \rotatebox{90}{Bicyclist} & \rotatebox{90}{Truck} & \rotatebox{90}{MBT} & \rotatebox{90}{Dog} & \rotatebox{90}{Wheelchair} & \rotatebox{90}{W-Device} & \rotatebox{90}{W-Ride} \\
        \midrule
            CenterPoint & 22.0 & 67.6 & 38.9 & 46.5 & 16.9 & 37.4 & 40.1 & 32.2 & 28.6 & 27.4 & 33.4 & 24.5 & 8.7 & 25.8 & 22.6 & 29.5 & 22.4 & 6.3 & 3.9 & 0.5 & 20.1 & 22.1 & 0.0 & 3.9 & 0.5 & 10.9 & 4.2 \\
            VoxelNeXt & 30.7 & 72.7 & 38.8 & 63.2 & 40.2 & 40.1 & 53.9 & 64.9 & 44.7 & 39.4 & 42.4 & 40.6 & 20.1 & 25.2 & 19.9 & 44.9 & 20.9 & 14.9 & 6.8 & 15.7 & 32.4 & 16.9 & 0.0 & 14.4 & 0.1 & 17.4 & 6.6 \\
            HEDNet & 37.1 & 78.2 & 47.7 & 67.6 & 46.4 & 45.9 & 56.9 & 67.0 & 48.7 & 46.5 & 58.2 & 47.5 & 23.3 & 40.9 & 27.5 & 46.8 & 27.9 & 20.6 & 6.9 & 27.2 & 38.7 & 21.6 & 0.0 & 30.7 & 9.5 & 28.5 & 8.7 \\
            SAFDNet & 39.7 & 78.5 & 49.4 & 70.7 & 51.5 & 44.7 & 65.7 & 72.3 & 54.3 & 49.7 & 60.8 & 50.0 & 31.3 & 44.9 & 24.7 & 55.4 & 31.4 & 22.1 & 7.1 & 31.1 & 42.7 & 23.6 & 0.0 & 26.1 & 1.4 & 30.2 & 11.5 \\
            
        \midrule
            SFMNet & \textbf{40.4} & 79.0 & 49.5 & 72.3 & 49.3 & 46.1 & 65.1 & 73.8 & 55.1 & 53.2 & 62.8 & 54.2 & 30.2 & 44.8 & 26.7 & 57.9 & 31.7 & 20.9 & 8.3 & 33.4 & 44.4 & 21.6 & 0.0 & 22.1 & 3.1 & 33.1 & 12.1 \\
         \bottomrule
    \end{tabular}
    \caption{
    \textbf{Results on the Argoverse2 validation set (long-range, up to $\pm 200$ meters).}
    The long-range detection dataset highlights our detector's ability to capture long-range dependencies.
    }
    \label{tab:argo2_results}

\vspace{0.02in}
    \scriptsize
    \setlength\tabcolsep{4pt} 
    \centering
    \begin{tabular}{c | l | c c | c c c c c c }
         \toprule
            \multirow{2}{*}{Design} & \multirow{2}{*}{Method} & \multicolumn{2}{c|}{mAP/mAPH} & \multicolumn{2}{c}{Vehicle AP/APH} & \multicolumn{2}{c}{Pedestrian AP/APH} & \multicolumn{2}{c}{Cyclist AP/APH} \\
             & & \multicolumn{1}{c}{L1} & \multicolumn{1}{c|}{L2} & \multicolumn{1}{c}{L1} & \multicolumn{1}{c}{L2} & \multicolumn{1}{c}{L1} & \multicolumn{1}{c}{L2} & \multicolumn{1}{c}{L1} & \multicolumn{1}{c}{L2} \\
        \midrule
            \multirow{6}{*}{Transformers} & SST~\cite{fan2022embracing} & 74.5/71.0 & 67.8/64.6 & 74.2/73.8 & 65.5/65.1 & 78.7/69.6 & 70.0/61.7 & 70.7/69.6 & 68.0/66.9 \\
            & TransFusion-L~\cite{bai2022transfusion} & -/- & -/64.9 & -/- & -/65.1 & -/- & -/63.7 & -/- & -/65.9 \\
            & SWFormer~\cite{sun2022swformer} & -/- & -/- & 77.8/77.3 & 69.2/68.8 & 80.9/72.7 & 72.5/64.9 & -/- & -/- \\
            & CenterFormer~\cite{zhou2022centerformer} & 75.6/73.2 & 71.4/69.1 & 75.0/74.4 & 69.9/69.4 & 78.0/72.4 & 73.1/67.7 & 73.8/72.7 & 71.3/70.2 \\
            & DSVT-Voxel~\cite{wang2023dsvt} & 80.3/78.2 & 74.0/72.1 & 79.7/79.3 & 71.4/71.0 & 83.7/78.9 & 76.1/71.5 & 77.5/76.5 & 74.6/73.7 \\
            & PTv3~\cite{wu2024point} & -/- & 73.0/70.5 & -/- & 71.2/70.8 & -/- & 76.3/70.4 & -/- & 71.5/70.4 \\
        \midrule
            \multirow{2}{*}{Large kernels} & LargeKernel3D~\cite{chen2023largekernel3d} & -/- & -/- & 78.1/77.6 & 69.8/69.4 & -/- & -/- & -/- & -/- \\
            & LinK~\cite{lu2023link} (6 epochs) & -/- & -/64.6 & -/- & -/65.0 & -/- & -/60.4 & -/- & -/68.4 \\
        \midrule
            \multirow{12}{*}{Small kernels} & SECOND~\cite{second} & 67.2/63.1 & 61.0/57.2 & 72.3/71.7 & 63.9/63.3 & 68.7/58.2 & 60.7/51.3 & 60.6/59.3 & 58.3/57.0 \\
            & PointPillars~\cite{pointpillars} & 69.0/63.5 & 62.8/57.8 & 72.1/71.5 & 63.6/63.1 & 70.6/56.7 & 62.8/50.3 & 64.4/62.3 & 61.9/59.9 \\
            & Part-A2~\cite{parta2} & 73.6/70.3 & 66.9/63.8 & 77.1/76.5 & 68.5/68.0 & 75.2/66.9 & 66.2/58.6 & 68.6/67.4 & 66.1/64.9 \\
            & PV-RCNN~\cite{pvrcnn} & 76.2/73.6 & 69.6/67.2 & 78.0/77.5 & 69.4/69.0 & 79.2/73.0 & 70.4/64.7 & 71.5/70.3 & 69.0/67.8 \\
            & CenterPoint~\cite{yin2021center} & 75.9/73.5 & 69.8/67.6 & 76.6/76.0 & 68.9/68.4 & 79.0/73.4 & 71.0/65.8 & 72.1/71.0 & 69.5/68.5\\
            & AFDetV2~\cite{hu2022afdetv2} & 77.2/74.8 & 71.0/68.8 & 77.6/77.1 & 69.7/69.2 & 80.2/74.6 & 72.2/67.0 & 73.7/72.7 & 71.0/70.1 \\
            & FSD~\cite{fan2022fully} & 79.6/77.4 & 72.9/70.8 & 79.2/78.8 & 70.5/70.1 & 82.6/77.3 & 73.9/69.1 & 77.1/76.0 & 74.4/73.3 \\
            & PV-RCNN++~\cite{pvrcnn++} & 78.1/75.9 & 71.7/69.5 & 79.3/78.8 & 70.6/70.2 & 81.3/76.3 & 73.2/68.0 & 73.7/72.7 & 71.2/70.2  \\
            & VoxelNeXt~\cite{chen2023voxelnext} & 78.6/76.3 & 72.2/70.1 & 78.2/77.7 & 69.9/69.4 & 81.5/76.3 & 73.5/68.6 & 76.1/74.9 & 73.3/72.2 \\
            & HEDNet~\cite{zhang2024hednet} & 81.4/79.4 & 75.3/73.4 & 81.1/80.6 & 73.2/72.7 & 84.4/80.0 & 76.8/72.6 & 78.7/77.7 & 75.8/74.9 \\
            & SAFDNet~\cite{zhang2024safdnet} & \textbf{81.8}/\textbf{79.9} & \textbf{75.7}/\textbf{73.9} & 80.6/80.1 & 72.7/72.3 & 84.7/80.4 & 77.3/73.1 & 80.0/79.0 & 77.2.76.2 \\

            \cmidrule{2-10}
            & SFMNet (ours) & \textbf{81.8}/\underline{79.8} & \textbf{75.7}/\underline{73.8} & 80.5/80.0 & 72.6/72.2 & 85.0/80.6 & 77.4/73.2 & 79.9/78.9 & 77.0/76.1 \\
         \bottomrule
    \end{tabular}
    \caption{
    \textbf{Results on the Waymo Open validation set (mid-range, up to $\pm 75$ meters).}
    Our SFMNet outperforms transformer- and large-kernel-based detectors, while achieving results comparable to the state-of-the-art small-kernel detector~\cite{zhang2024safdnet}.
    }
    \label{tab:WOD_results}

\vspace{0.02in}
    \scriptsize
    \setlength\tabcolsep{4pt} 
    \centering
    \begin{tabular}{c | l | c c | c c c c c c c c c c }
         \toprule
            Design & Method & NDS & mAP & Car & Truck & Bus & Trailer & C.V. & Pedestrian & Motor & Bike & Cone & Barrier \\
        \midrule
            \multirow{2}{*}{Transformers} & TransFusion-L~\cite{bai2022transfusion} & 70.1 & 65.5 & 86.9 & 60.8 & 73.1 & 43.4 & 25.2 & 87.5 & 72.9 & 57.3 & 77.2 & 70.3 \\
            & DSVT-Voxel~\cite{wang2023dsvt} & 71.1 & 66.4 & 87.4 & 62.6 & 75.9 & 42.1 & 25.3 & 88.2 & 74.8 & 58.7 & 77.8 & 70.9 \\
        \midrule
            \multirow{2}{*}{Large kernels} & LargeKernel3D~\cite{chen2023largekernel3d} & 69.1 & 63.9 & 85.1 & 60.1 & 72.6 & 41.4 & 24.3 & 85.6 & 70.8 & 59.2 & 72.3 & 67.7 \\
            & LinK~\cite{lu2023link} & 69.5 & 63.6 & - & - & - & - & - & - & - & - & - & - \\
        \midrule
            \multirow{6}{*}{Small kernels} & CenterPoint~\cite{yin2021center} & 66.5 & 59.2 & 84.9 & 57.4 & 70.7 & 38.1 & 16.9 & 85.1 & 59.0 & 42.0 & 69.8 & 68.3 \\
            & VoxelNeXt~\cite{chen2023voxelnext} & 66.7 & 60.5 & 83.9 & 55.5 & 70.5 & 38.1 & 21.1 & 84.6 & 62.8 & 50.0 & 69.4 & 69.4 \\
            & PillarNeXt-L~\cite{li2023pillarnext} & 68.4 & 62.2 & 85.0 & 57.4 & 67.6 & 35.6 & 20.6 & 86.8 & 68.6 & 53.1 & 77.3 & 69.7 \\
            & HEDNet~\cite{zhang2024hednet} & \textbf{71.4} & \underline{66.7} & 87.7 & 60.6 & 77.8 & 50.7 & 28.9 & 87.1 & 74.3 & 56.8 & 76.3 & 66.9 \\
            & SAFDNet~\cite{zhang2024safdnet} & 71.0 & 66.3 & 87.6 & 60.8 & 78.0 & 43.5 & 26.6 & 87.8 & 75.5 & 58.0 & 75.0 & 69.7 \\
        
        \cmidrule{2-14}
            & SFMNet (ours) & \underline{71.3} & \textbf{66.9} & 87.6 & 63.2 & 76.2 & 48.9 & 26.0 & 88.4 & 74.7 & 56.9 & 78.5 & 69.1 \\
         \bottomrule
    \end{tabular}
    \caption{
    \textbf{Results on the nuScenes validation set (short-range, up to $\pm 54$ meters).}
    Our SFMNet outperforms all transformer- and large-kernel-based detectors, surpassing the previous SoTA small-kernel detector on the mAP metric and ranking second on the NDS metric. 
    }
    \label{tab:nuscenes_results}
\end{table*}

\noindent
\textbf{Implementations details.}
To compare with prior state-of-the-art methods, we trained SFMNet for $24$ epochs on Argoverse2 and Waymo Open, and $20$ epochs on nuScenes.
Ablation studies used 6 epochs on Argoverse2.
Additional implementation details are provided in the supplemental.

\subsection{Results}

\textbf{Quantitative results.} 
We expect our model to be particularly beneficial for long-range scenes.
Indeed, as shown in \tabref{tab:argo2_results}, which compares our method to state-of-the-art sparse convolution-based detectors with small kernels, our detector achieves a $0.7\%$ mAP improvement over the previous best method~\cite{zhang2024safdnet}, demonstrating its ability to capture long-range dependencies. 
For mid-range detection on Waymo Open, \tabref{tab:WOD_results} compares our method to state-of-the-art methods on the validation set, where '-' indicates unavailable results. 
Our detector outperforms both transformer-based and large kernel-based methods, achieving gains of $1.7\%$ and $3.3\%$ mAPH on Level-2 difficulty compared to the transformer-based detectors DSVT~\cite{wang2023dsvt} and PTv3~\cite{wu2024point}, respectively, while also achieving comparable results to the best small kernel-based detector~\cite{zhang2024safdnet}. 
For the short-range detection dataset nuScenes,\tabref{tab:nuscenes_results} shows that our method achieves SoTA results on the mAP metric and ranks second on the NDS metric, outperforming both large kernel and transformer-based methods.
The effective receptive field (ERF) is $3.3$ meters for Argoverse2 and $2$ meters for Waymo Open and nuScenes.

\begin{figure*}
    \setlength\tabcolsep{3pt} 
    \centering
    \begin{tabular}{ccc}
    \includegraphics[width=0.31\linewidth]{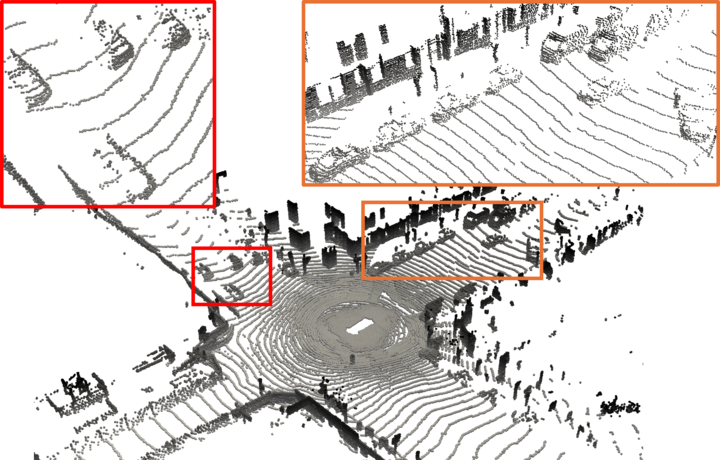} &
    \includegraphics[width=0.31\linewidth]{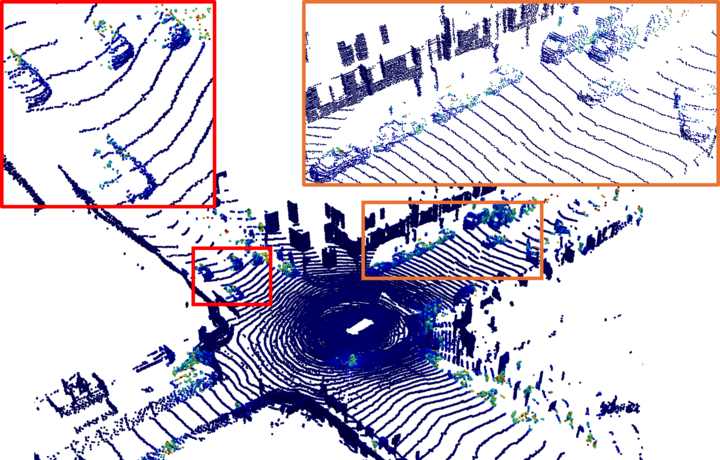} &
    \includegraphics[width=0.31\linewidth]{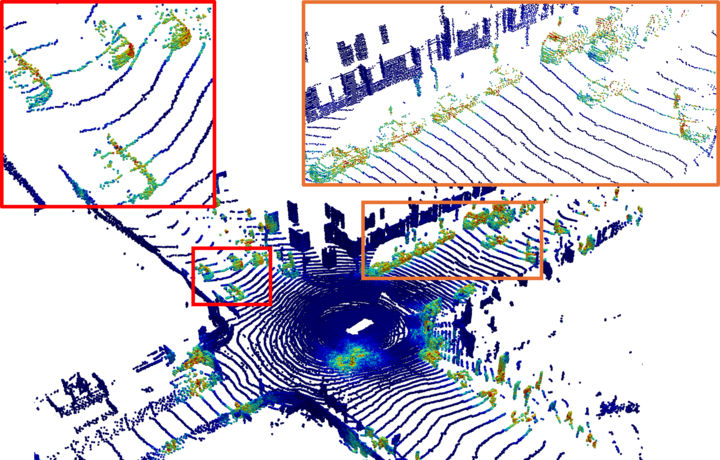} \\
    \small (a) Input scene & \small (b) ERF without SFM  & \small (c) ERF with SFM
    \end{tabular}
    \caption{
    \textbf{Effective receptive field (ERF) of SFMNet.}
    (a) A typical input scene with height is shown in grayscale.
    (b,c) The ERF for each ground-truth object is color-coded from blue (outside ERF) to red (high-attention regions). 
    Red and orange rectangles highlight zoomed-in regions.
    In the right (orange) zoom-in, four stationary cars and a van, plus two moving cars and a van, are shown. 
    Without the SFM module (b), only a local portion of each vehicle is within the ERF. 
    With SFM (c), the entire object is captured, providing more context and spatial understanding for accurate 3D detection.
    }
    \label{fig:ERF}
\end{figure*}

\noindent
\textbf{Qualitative results.}
To qualitatively validate our detector's ability to capture an extensive ERF, for each ground-truth object, we calculate the ERF by sampling a query voxel and analyzing its reach.
\figref{fig:ERF} visualizes the ERF (color-coded from blue to red) of our detector with and without our SFM module.
The more yellowish the color, the larger the ERF. 
For each ground-truth object, we randomly select one voxel to serve as the query and calculate the ERF associated with it.
Indeed, the dependency coverage leads to large regional receptive fields.
Without the SFM module (b), the receptive field is limited to local areas around each query, capturing only fragments of the objects. 
For instance, in the left zoom-in in \figref{fig:ERF}(b), there are four vehicles. 
The receptive field covers only portions of each vehicle, leaving most voxels outside this field (blue).
In contrast, with our SFM module (c), the detector gains a broader receptive field, allowing it to capture even partially occluded vehicles.
This is reflected in the four vehicles being highlighted in green, yellow, and red colors.

\begin{table}[t]
    \centering
        \small
        \setlength\tabcolsep{3pt} 
        \centering
        \begin{tabular}{l | c c | c }
            \toprule
                Method & Window(m) & Memory(GB) & mAP \\
            \midrule
                Global-attention & 400 & OOM & - \\
                Local-attention & 3.3 & 44.06 & 34.8 \\
                Set-attention (\cite{wang2023dsvt}) & 3.6 & 44.21 & 35.5 \\
            \midrule
                SFM (ours) & 1.9 & 32.76 & 36.8 \\
                SFM (ours) & 3.3 & 37.53 & 37.1 \\
            \bottomrule
        \end{tabular}
        \caption{
        \textbf{Comparison to attention.}
        While global attention covers the entire ERF, it quickly exceeds memory limits. 
        In contrast, Local-attention and Set-attention~\cite{wang2023dsvt} focus on voxels within a local context window. 
        Our SFM module outperforms both in accuracy and efficiency, even with a smaller window.
        }
        \label{tab:sfm_ablation_argo2_atten}
\end{table}

\begin{table}[]
        \footnotesize
        \setlength\tabcolsep{4pt} 
        \centering
        \begin{tabular}{c | c c c c c | c}
            \toprule
                & \multicolumn{2}{c}{Stage 4} & \multicolumn{2}{c}{2D backbone} & \multirow{2}{*}{\#Param.(M)} & \multirow{2}{*}{mAP} \\
                & \#SRB & Multi. & \#SRB & Multi. & & \\
            \midrule
                \multirow{3}{*}{A} & 2 & 1 & 4 & 1 & 3.84 & 36.7 \\
                & 4 & 1 & 4 & 1 & 4.29 & 37.2 \\
                & 6 & 1 & 4 & 1 & 4.73 & 37.5 \\
            \midrule
                \multirow{3}{*}{B} & 2 & 2 & 4 & 2 & 5.27 & 38.0 \\
                & 2 & 4 & 4 & 2 & 7.14 & 38.6 \\
                & 2 & 6 & 4 & 2 & 9.01 & 39.1 \\
            \midrule
                C & 4 & 2 & 4 & 2 & 6.16 & 38.9 \\
            \bottomrule
        \end{tabular}
        \caption{
        \textbf{Effect of SFM blocks \& SRBs.}
        Stacking SRBs after each SFM block improves accuracy (A). 
        Increasing SFM repetitions boosts it further (B), but with more parameters. 
        Configuration (C) offers a balanced trade-off.
        }
        \label{tab:sfm_ablation_argo2_blocks}
\end{table}

\subsection{Ablation study}
\label{sec:ablation}

\tabref{tab:sfm_ablation_argo2_atten} compares our SFM module to traditional global and local attention, as well as set attention~\cite{wang2023dsvt}, on the Argoverse2 dataset over six epochs.
For this evaluation, we replaced an attention block in the 3D and 2D backbones with either a single SFM module or an alternative attention mechanism, demonstrating that our SFM achieves higher accuracy and efficiency.
While global attention attains a high ERF, it quickly reaches memory limits ($80$ GB) due to the large voxel count.
In contrast, local and set attention operate within a limited context window, with set attention outperforming local attention.
However, our SFM module surpasses both in accuracy (mAP) and efficiency (memory usage) despite using a smaller context window.

\tabref{tab:sfm_ablation_argo2_blocks} examines the effect of varying the number of SFM blocks and SRBs.
Increasing the number of SRBs after the SFM block in Stage 4 (A) improves accuracy with a modest parameter increase.
Similarly, adding more SFM block repetitions (Multipliers) with SRBs (B) boosts accuracy but at a higher parameter cost.
A balanced approach, with moderate repetitions and additional SRBs (C), provides a good trade-off between accuracy and parameters.

\begin{table}[t]
    \small
    \setlength\tabcolsep{4pt} 
    \centering
    \begin{tabular}{c c c c | c }
        \toprule
            \#levels & dilation rates & ERF(m) & Memory(GB) & mAP \\
        \midrule
            2 & [1,3] & 0.9 & 30.19 & 35.7 \\
            3 & [1,3,5] & 1.9 & 32.76 & 36.8 \\
            3 & [1,5,9] & 3.1 & 32.98 & 36.4 \\
            4 & [1,3,5,7] & 3.3 & 37.53 & 37.1 \\
            5 & [1,3,5,7,9] & 5.1 & 36.39 & 37.1 \\
        \bottomrule
    \end{tabular}
    \caption{
    \textbf{ERF effects.}
    We evaluate different ERF sizes by increasing the number of context levels, each using a sparse convolution with kernel size $3$ and varying dilation. 
    Four levels, yielding an ERF of $3.3$ meters, achieve the best accuracy.
    }
    \label{tab:ablation_argo2_ctx_levels}
\end{table}

\tabref{tab:ablation_argo2_ctx_levels} investigates the impact of different context levels with varying ERF. 
Increasing levels improves accuracy but also raises memory usage. 
Beyond a point, more levels do not help, as the network tends to ignore overly long-range features. 
Relying only on high dilation rates without local contexts (third row) yields suboptimal performance, highlighting the need for short-range dependencies. 
Additional ablations on SFM placement are in the supplemental.



\noindent
\textbf{Limitations.}
Our method is designed to capture long-range dependencies and enhance models that lack this capability. On datasets requiring long-range reasoning, it improves not only those models but also others with a similar objective. However, on short-range data, while it still benefits the former, its performance is on par with the latter.

\section{Conclusion}
This paper introduced a novel 3D Focal Modulation module, SFM, designed to effectively capture dependencies from short- to long-range. 
It is based on a hierarchical submanifold sparse convolution design to efficiently handle data sparsity.
It also presented SFMNet, a new sparse convolution-based 3D detector built on top of the SFM.
It addresses the limitation of traditional sparse convolution detectors by significantly enlarging the effective receptive field. 
Experimental results demonstrate that SFMNet captures dependencies very well and achieves SoTA performance on multi-range datasets.

\clearpage
{
    \small
    \bibliographystyle{ieeenat_fullname}
    \bibliography{references}

@inproceedings{pvrcnn,
  title={Pv-rcnn: Point-voxel feature set abstraction for 3d object detection},
  author={Shi, Shaoshuai and Guo, Chaoxu and Jiang, Li and Wang, Zhe and Shi, Jianping and Wang, Xiaogang and Li, Hongsheng},
  booktitle={Proceedings of the IEEE/CVF Conference on Computer Vision and Pattern Recognition},
  pages={10529--10538},
  year={2020}
}

@inproceedings{pointpillars,
  title={Pointpillars: Fast encoders for object detection from point clouds},
  author={Lang, Alex H and Vora, Sourabh and Caesar, Holger and Zhou, Lubing and Yang, Jiong and Beijbom, Oscar},
  booktitle={Proceedings of the IEEE/CVF Conference on Computer Vision and Pattern Recognition},
  pages={12697--12705},
  year={2019}
}

@article{voxelrcnn,
  title={Voxel r-cnn: Towards high performance voxel-based 3d object detection},
  author={Deng, Jiajun and Shi, Shaoshuai and Li, Peiwei and Zhou, Wengang and Zhang, Yanyong and Li, Houqiang},
  journal={arXiv preprint arXiv:2012.15712},
  volume={1},
  number={2},
  pages={4},
  year={2020}
}

@article{second,
  title={Second: Sparsely embedded convolutional detection},
  author={Yan, Yan and Mao, Yuxing and Li, Bo},
  journal={Sensors},
  volume={18},
  number={10},
  pages={3337},
  year={2018},
  publisher={Multidisciplinary Digital Publishing Institute}
}

@inproceedings{pointrcnn,
  title={Pointrcnn: 3d object proposal generation and detection from point cloud},
  author={Shi, Shaoshuai and Wang, Xiaogang and Li, Hongsheng},
  booktitle={Proceedings of the IEEE/CVF conference on computer vision and pattern recognition},
  pages={770--779},
  year={2019}
}

@inproceedings{zhou2018voxelnet,
  title={Voxelnet: End-to-end learning for point cloud based 3d object detection},
  author={Zhou, Yin and Tuzel, Oncel},
  booktitle={Proceedings of the IEEE conference on computer vision and pattern recognition},
  pages={4490--4499},
  year={2018}
}

@inproceedings{sparseconv,
  title={3d semantic segmentation with submanifold sparse convolutional networks},
  author={Graham, Benjamin and Engelcke, Martin and Van Der Maaten, Laurens},
  booktitle={Proceedings of the IEEE conference on computer vision and pattern recognition},
  pages={9224--9232},
  year={2018}
}

@inproceedings{chen2019fast,
  title={Fast point r-cnn},
  author={Chen, Yilun and Liu, Shu and Shen, Xiaoyong and Jia, Jiaya},
  booktitle={Proceedings of the IEEE/CVF International Conference on Computer Vision},
  pages={9775--9784},
  year={2019}
}

@inproceedings{qi2018frustum,
  title={Frustum pointnets for 3d object detection from rgb-d data},
  author={Qi, Charles R and Liu, Wei and Wu, Chenxia and Su, Hao and Guibas, Leonidas J},
  booktitle={Proceedings of the IEEE conference on computer vision and pattern recognition},
  pages={918--927},
  year={2018}
}

@inproceedings{qi2017pointnet,
  title={Pointnet: Deep learning on point sets for 3d classification and segmentation},
  author={Qi, Charles R and Su, Hao and Mo, Kaichun and Guibas, Leonidas J},
  booktitle={Proceedings of the IEEE conference on computer vision and pattern recognition},
  pages={652--660},
  year={2017}
}

@article{qi2017pointnet++,
  title={Pointnet++: Deep hierarchical feature learning on point sets in a metric space},
  author={Qi, Charles Ruizhongtai and Yi, Li and Su, Hao and Guibas, Leonidas J},
  journal={Advances in neural information processing systems},
  volume={30},
  year={2017}
}

@inproceedings{yang20203dssd,
  title={3dssd: Point-based 3d single stage object detector},
  author={Yang, Zetong and Sun, Yanan and Liu, Shu and Jia, Jiaya},
  booktitle={Proceedings of the IEEE/CVF conference on computer vision and pattern recognition},
  pages={11040--11048},
  year={2020}
}

@inproceedings{ye2020hvnet,
  title={Hvnet: Hybrid voxel network for lidar based 3d object detection},
  author={Ye, Maosheng and Xu, Shuangjie and Cao, Tongyi},
  booktitle={Proceedings of the IEEE/CVF conference on computer vision and pattern recognition},
  pages={1631--1640},
  year={2020}
}

@inproceedings{kitti1,
  title={Are we ready for autonomous driving? the kitti vision benchmark suite},
  author={Geiger, Andreas and Lenz, Philip and Urtasun, Raquel},
  booktitle={2012 IEEE conference on computer vision and pattern recognition},
  pages={3354--3361},
  year={2012},
  organization={IEEE}
}

@inproceedings{yin2021center,
  title={Center-based 3d object detection and tracking},
  author={Yin, Tianwei and Zhou, Xingyi and Krahenbuhl, Philipp},
  booktitle={Proceedings of the IEEE/CVF conference on computer vision and pattern recognition},
  pages={11784--11793},
  year={2021}
}

@misc{openpcdet2020,
    title={OpenPCDet: An Open-source Toolbox for 3D Object Detection from Point Clouds},
    author={OpenPCDet Development Team},
    howpublished = {\url{https://github.com/open-mmlab/OpenPCDet}},
    year={2020}
}

@article{parta2,
  title={From points to parts: 3d object detection from point cloud with part-aware and part-aggregation network},
  author={Shi, Shaoshuai and Wang, Zhe and Shi, Jianping and Wang, Xiaogang and Li, Hongsheng},
  journal={IEEE transactions on pattern analysis and machine intelligence},
  volume={43},
  number={8},
  pages={2647--2664},
  year={2020},
  publisher={IEEE}
}

@article{graham2017submanifold,
  title={Submanifold sparse convolutional networks},
  author={Graham, Benjamin and van der Maaten, Laurens},
  journal={arXiv preprint arXiv:1706.01307},
  year={2017}
}

@article{thrun2006stanley,
  title={Stanley: The robot that won the DARPA Grand Challenge},
  author={Thrun, Sebastian and Montemerlo, Mike and Dahlkamp, Hendrik and Stavens, David and Aron, Andrei and Diebel, James and Fong, Philip and Gale, John and Halpenny, Morgan and Hoffmann, Gabriel and others},
  journal={Journal of field Robotics},
  volume={23},
  number={9},
  pages={661--692},
  year={2006},
  publisher={Wiley Online Library}
}

@inproceedings{zheng2021se,
  title={SE-SSD: Self-ensembling single-stage object detector from point cloud},
  author={Zheng, Wu and Tang, Weiliang and Jiang, Li and Fu, Chi-Wing},
  booktitle={Proceedings of the IEEE/CVF Conference on Computer Vision and Pattern Recognition},
  pages={14494--14503},
  year={2021}
}

@inproceedings{zheng2021cia,
  title={Cia-ssd: Confident iou-aware single-stage object detector from point cloud},
  author={Zheng, Wu and Tang, Weiliang and Chen, Sijin and Jiang, Li and Fu, Chi-Wing},
  booktitle={Proceedings of the AAAI conference on artificial intelligence},
  volume={35},
  number={4},
  pages={3555--3562},
  year={2021}
}

@inproceedings{shi2020point,
  title={Point-gnn: Graph neural network for 3d object detection in a point cloud},
  author={Shi, Weijing and Rajkumar, Raj},
  booktitle={Proceedings of the IEEE/CVF conference on computer vision and pattern recognition},
  pages={1711--1719},
  year={2020}
}

@inproceedings{waymo,
  title={Scalability in perception for autonomous driving: Waymo open dataset},
  author={Sun, Pei and Kretzschmar, Henrik and Dotiwalla, Xerxes and Chouard, Aurelien and Patnaik, Vijaysai and Tsui, Paul and Guo, James and Zhou, Yin and Chai, Yuning and Caine, Benjamin and others},
  booktitle={Proceedings of the IEEE/CVF conference on computer vision and pattern recognition},
  pages={2446--2454},
  year={2020}
}

@article{pvrcnn++,
  title={PV-RCNN++: Point-voxel feature set abstraction with local vector representation for 3D object detection},
  author={Shi, Shaoshuai and Jiang, Li and Deng, Jiajun and Wang, Zhe and Guo, Chaoxu and Shi, Jianping and Wang, Xiaogang and Li, Hongsheng},
  journal={International Journal of Computer Vision},
  volume={131},
  number={2},
  pages={531--551},
  year={2023},
  publisher={Springer}
}

@article{once,
  title={One million scenes for autonomous driving: Once dataset},
  author={Mao, Jiageng and Niu, Minzhe and Jiang, Chenhan and Liang, Hanxue and Chen, Jingheng and Liang, Xiaodan and Li, Yamin and Ye, Chaoqiang and Zhang, Wei and Li, Zhenguo and others},
  journal={arXiv preprint arXiv:2106.11037},
  year={2021}
}

@inproceedings{nuscenes,
  title={nuscenes: A multimodal dataset for autonomous driving},
  author={Caesar, Holger and Bankiti, Varun and Lang, Alex H and Vora, Sourabh and Liong, Venice Erin and Xu, Qiang and Krishnan, Anush and Pan, Yu and Baldan, Giancarlo and Beijbom, Oscar},
  booktitle={Proceedings of the IEEE/CVF conference on computer vision and pattern recognition},
  pages={11621--11631},
  year={2020}
}

@inproceedings{shrout2023gravos,
  title={GraVoS: Voxel Selection for 3D Point-Cloud Detection},
  author={Shrout, Oren and Ben-Shabat, Yizhak and Tal, Ayellet},
  booktitle={Proceedings of the IEEE/CVF Conference on Computer Vision and Pattern Recognition},
  pages={21684--21693},
  year={2023}
}

@inproceedings{chang2019argoverse,
  title={Argoverse: 3d tracking and forecasting with rich maps},
  author={Chang, Ming-Fang and Lambert, John and Sangkloy, Patsorn and Singh, Jagjeet and Bak, Slawomir and Hartnett, Andrew and Wang, De and Carr, Peter and Lucey, Simon and Ramanan, Deva and others},
  booktitle={Proceedings of the IEEE/CVF conference on computer vision and pattern recognition},
  pages={8748--8757},
  year={2019}
}

@article{wilson2023argoverse,
  title={Argoverse 2: Next generation datasets for self-driving perception and forecasting},
  author={Wilson, Benjamin and Qi, William and Agarwal, Tanmay and Lambert, John and Singh, Jagjeet and Khandelwal, Siddhesh and Pan, Bowen and Kumar, Ratnesh and Hartnett, Andrew and Pontes, Jhony Kaesemodel and others},
  journal={arXiv preprint arXiv:2301.00493},
  year={2023}
}

@article{zhang2024hednet,
  title={Hednet: A hierarchical encoder-decoder network for 3d object detection in point clouds},
  author={Zhang, Gang and Junnan, Chen and Gao, Guohuan and Li, Jianmin and Hu, Xiaolin},
  journal={Advances in Neural Information Processing Systems},
  volume={36},
  year={2024}
}

@inproceedings{chen2023voxelnext,
  title={Voxelnext: Fully sparse voxelnet for 3d object detection and tracking},
  author={Chen, Yukang and Liu, Jianhui and Zhang, Xiangyu and Qi, Xiaojuan and Jia, Jiaya},
  booktitle={Proceedings of the IEEE/CVF Conference on Computer Vision and Pattern Recognition},
  pages={21674--21683},
  year={2023}
}

@inproceedings{wang2023dsvt,
  title={Dsvt: Dynamic sparse voxel transformer with rotated sets},
  author={Wang, Haiyang and Shi, Chen and Shi, Shaoshuai and Lei, Meng and Wang, Sen and He, Di and Schiele, Bernt and Wang, Liwei},
  booktitle={Proceedings of the IEEE/CVF Conference on Computer Vision and Pattern Recognition},
  pages={13520--13529},
  year={2023}
}

@inproceedings{wang2018non,
  title={Non-local neural networks},
  author={Wang, Xiaolong and Girshick, Ross and Gupta, Abhinav and He, Kaiming},
  booktitle={Proceedings of the IEEE conference on computer vision and pattern recognition},
  pages={7794--7803},
  year={2018}
}

@inproceedings{zhu2017target,
  title={Target-driven visual navigation in indoor scenes using deep reinforcement learning},
  author={Zhu, Yuke and Mottaghi, Roozbeh and Kolve, Eric and Lim, Joseph J and Gupta, Abhinav and Fei-Fei, Li and Farhadi, Ali},
  booktitle={2017 IEEE international conference on robotics and automation (ICRA)},
  pages={3357--3364},
  year={2017},
  organization={IEEE}
}

@article{tian2022fully,
  title={Fully convolutional one-stage 3d object detection on lidar range images},
  author={Tian, Zhi and Chu, Xiangxiang and Wang, Xiaoming and Wei, Xiaolin and Shen, Chunhua},
  journal={Advances in Neural Information Processing Systems},
  volume={35},
  pages={34899--34911},
  year={2022}
}

@inproceedings{sun2021rsn,
  title={Rsn: Range sparse net for efficient, accurate lidar 3d object detection},
  author={Sun, Pei and Wang, Weiyue and Chai, Yuning and Elsayed, Gamaleldin and Bewley, Alex and Zhang, Xiao and Sminchisescu, Cristian and Anguelov, Dragomir},
  booktitle={Proceedings of the IEEE/CVF Conference on Computer Vision and Pattern Recognition},
  pages={5725--5734},
  year={2021}
}

@inproceedings{fan2021rangedet,
  title={Rangedet: In defense of range view for lidar-based 3d object detection},
  author={Fan, Lue and Xiong, Xuan and Wang, Feng and Wang, Naiyan and Zhang, Zhaoxiang},
  booktitle={Proceedings of the IEEE/CVF international conference on computer vision},
  pages={2918--2927},
  year={2021}
}

@inproceedings{zhang2024safdnet,
  title={SAFDNet: A Simple and Effective Network for Fully Sparse 3D Object Detection},
  author={Zhang, Gang and Chen, Junnan and Gao, Guohuan and Li, Jianmin and Liu, Si and Hu, Xiaolin},
  booktitle={Proceedings of the IEEE/CVF Conference on Computer Vision and Pattern Recognition},
  pages={14477--14486},
  year={2024}
}

@article{graham2015sparse,
  title={Sparse 3D convolutional neural networks},
  author={Graham, Ben},
  journal={arXiv preprint arXiv:1505.02890},
  year={2015}
}

@inproceedings{choy20194d,
  title={4d spatio-temporal convnets: Minkowski convolutional neural networks},
  author={Choy, Christopher and Gwak, JunYoung and Savarese, Silvio},
  booktitle={Proceedings of the IEEE/CVF conference on computer vision and pattern recognition},
  pages={3075--3084},
  year={2019}
}

@inproceedings{fan2022embracing,
  title={Embracing single stride 3d object detector with sparse transformer},
  author={Fan, Lue and Pang, Ziqi and Zhang, Tianyuan and Wang, Yu-Xiong and Zhao, Hang and Wang, Feng and Wang, Naiyan and Zhang, Zhaoxiang},
  booktitle={Proceedings of the IEEE/CVF conference on computer vision and pattern recognition},
  pages={8458--8468},
  year={2022}
}

@inproceedings{zhou2022centerformer,
  title={Centerformer: Center-based transformer for 3d object detection},
  author={Zhou, Zixiang and Zhao, Xiangchen and Wang, Yu and Wang, Panqu and Foroosh, Hassan},
  booktitle={European Conference on Computer Vision},
  pages={496--513},
  year={2022},
  organization={Springer}
}

@inproceedings{sun2022swformer,
  title={Swformer: Sparse window transformer for 3d object detection in point clouds},
  author={Sun, Pei and Tan, Mingxing and Wang, Weiyue and Liu, Chenxi and Xia, Fei and Leng, Zhaoqi and Anguelov, Dragomir},
  booktitle={European Conference on Computer Vision},
  pages={426--442},
  year={2022},
  organization={Springer}
}

@inproceedings{chen2023focalformer3d,
  title={Focalformer3d: focusing on hard instance for 3d object detection},
  author={Chen, Yilun and Yu, Zhiding and Chen, Yukang and Lan, Shiyi and Anandkumar, Anima and Jia, Jiaya and Alvarez, Jose M},
  booktitle={Proceedings of the IEEE/CVF International Conference on Computer Vision},
  pages={8394--8405},
  year={2023}
}

@inproceedings{hu2022afdetv2,
  title={Afdetv2: Rethinking the necessity of the second stage for object detection from point clouds},
  author={Hu, Yihan and Ding, Zhuangzhuang and Ge, Runzhou and Shao, Wenxin and Huang, Li and Li, Kun and Liu, Qiang},
  booktitle={Proceedings of the AAAI Conference on Artificial Intelligence},
  volume={36},
  number={1},
  pages={969--979},
  year={2022}
}

@inproceedings{chen2023largekernel3d,
  title={Largekernel3d: Scaling up kernels in 3d sparse cnns},
  author={Chen, Yukang and Liu, Jianhui and Zhang, Xiangyu and Qi, Xiaojuan and Jia, Jiaya},
  booktitle={Proceedings of the IEEE/CVF Conference on Computer Vision and Pattern Recognition},
  pages={13488--13498},
  year={2023}
}

@article{fan2022fully,
  title={Fully sparse 3d object detection},
  author={Fan, Lue and Wang, Feng and Wang, Naiyan and Zhang, Zhao-Xiang},
  journal={Advances in Neural Information Processing Systems},
  volume={35},
  pages={351--363},
  year={2022}
}

@inproceedings{wu2024point,
  title={Point Transformer V3: Simpler Faster Stronger},
  author={Wu, Xiaoyang and Jiang, Li and Wang, Peng-Shuai and Liu, Zhijian and Liu, Xihui and Qiao, Yu and Ouyang, Wanli and He, Tong and Zhao, Hengshuang},
  booktitle={Proceedings of the IEEE/CVF Conference on Computer Vision and Pattern Recognition},
  pages={4840--4851},
  year={2024}
}

@inproceedings{bai2022transfusion,
  title={Transfusion: Robust lidar-camera fusion for 3d object detection with transformers},
  author={Bai, Xuyang and Hu, Zeyu and Zhu, Xinge and Huang, Qingqiu and Chen, Yilun and Fu, Hongbo and Tai, Chiew-Lan},
  booktitle={Proceedings of the IEEE/CVF conference on computer vision and pattern recognition},
  pages={1090--1099},
  year={2022}
}

@inproceedings{li2023pillarnext,
  title={PillarNeXt: Rethinking network designs for 3D object detection in LiDAR point clouds},
  author={Li, Jinyu and Luo, Chenxu and Yang, Xiaodong},
  booktitle={Proceedings of the IEEE/CVF Conference on Computer Vision and Pattern Recognition},
  pages={17567--17576},
  year={2023}
}

@inproceedings{wang2023unitr,
  title={Unitr: A unified and efficient multi-modal transformer for bird's-eye-view representation},
  author={Wang, Haiyang and Tang, Hao and Shi, Shaoshuai and Li, Aoxue and Li, Zhenguo and Schiele, Bernt and Wang, Liwei},
  booktitle={Proceedings of the IEEE/CVF International Conference on Computer Vision},
  pages={6792--6802},
  year={2023}
}

@inproceedings{lu2023link,
  title={Link: Linear kernel for lidar-based 3d perception},
  author={Lu, Tao and Ding, Xiang and Liu, Haisong and Wu, Gangshan and Wang, Limin},
  booktitle={Proceedings of the IEEE/CVF Conference on Computer Vision and Pattern Recognition},
  pages={1105--1115},
  year={2023}
}

@inproceedings{feng2024lsk3dnet,
  title={LSK3DNet: Towards Effective and Efficient 3D Perception with Large Sparse Kernels},
  author={Feng, Tuo and Wang, Wenguan and Ma, Fan and Yang, Yi},
  booktitle={Proceedings of the IEEE/CVF Conference on Computer Vision and Pattern Recognition},
  pages={14916--14927},
  year={2024}
}

@inproceedings{ding2022scaling,
  title={Scaling up your kernels to 31x31: Revisiting large kernel design in cnns},
  author={Ding, Xiaohan and Zhang, Xiangyu and Han, Jungong and Ding, Guiguang},
  booktitle={Proceedings of the IEEE/CVF conference on computer vision and pattern recognition},
  pages={11963--11975},
  year={2022}
}

@inproceedings{ding2024unireplknet,
  title={UniRepLKNet: A Universal Perception Large-Kernel ConvNet for Audio Video Point Cloud Time-Series and Image Recognition},
  author={Ding, Xiaohan and Zhang, Yiyuan and Ge, Yixiao and Zhao, Sijie and Song, Lin and Yue, Xiangyu and Shan, Ying},
  booktitle={Proceedings of the IEEE/CVF Conference on Computer Vision and Pattern Recognition},
  pages={5513--5524},
  year={2024}
}

@article{yang2022focal,
  title={Focal modulation networks},
  author={Yang, Jianwei and Li, Chunyuan and Dai, Xiyang and Gao, Jianfeng},
  journal={Advances in Neural Information Processing Systems},
  volume={35},
  pages={4203--4217},
  year={2022}
}

@article{vaswani2017attention,
  title={Attention is all you need},
  author={Vaswani, A},
  journal={Advances in Neural Information Processing Systems},
  year={2017}
}

@article{hendrycks2016gaussian,
  title={Gaussian error linear units (gelus)},
  author={Hendrycks, Dan and Gimpel, Kevin},
  journal={arXiv preprint arXiv:1606.08415},
  year={2016}
}

@inproceedings{zhou2020end,
  title={End-to-end multi-view fusion for 3d object detection in lidar point clouds},
  author={Zhou, Yin and Sun, Pei and Zhang, Yu and Anguelov, Dragomir and Gao, Jiyang and Ouyang, Tom and Guo, James and Ngiam, Jiquan and Vasudevan, Vijay},
  booktitle={Conference on Robot Learning},
  pages={923--932},
  year={2020},
  organization={PMLR}
}

@inproceedings{yu2022metaformer,
  title={Metaformer is actually what you need for vision},
  author={Yu, Weihao and Luo, Mi and Zhou, Pan and Si, Chenyang and Zhou, Yichen and Wang, Xinchao and Feng, Jiashi and Yan, Shuicheng},
  booktitle={Proceedings of the IEEE/CVF conference on computer vision and pattern recognition},
  pages={10819--10829},
  year={2022}
}

@article{liu2022more,
  title={More convnets in the 2020s: Scaling up kernels beyond 51x51 using sparsity},
  author={Liu, Shiwei and Chen, Tianlong and Chen, Xiaohan and Chen, Xuxi and Xiao, Qiao and Wu, Boqian and K{\"a}rkk{\"a}inen, Tommi and Pechenizkiy, Mykola and Mocanu, Decebal and Wang, Zhangyang},
  journal={arXiv preprint arXiv:2207.03620},
  year={2022}
}

@article{chen2024revealing,
  title={Revealing the Dark Secrets of Extremely Large Kernel ConvNets on Robustness},
  author={Chen, Honghao and Zhang, Yurong and Feng, Xiaokun and Chu, Xiangxiang and Huang, Kaiqi},
  journal={arXiv preprint arXiv:2407.08972},
  year={2024}
}

@inproceedings{pan20213d,
  title={3d object detection with pointformer},
  author={Pan, Xuran and Xia, Zhuofan and Song, Shiji and Li, Li Erran and Huang, Gao},
  booktitle={Proceedings of the IEEE/CVF conference on computer vision and pattern recognition},
  pages={7463--7472},
  year={2021}
}

@inproceedings{zhao2021point,
  title={Point transformer},
  author={Zhao, Hengshuang and Jiang, Li and Jia, Jiaya and Torr, Philip HS and Koltun, Vladlen},
  booktitle={Proceedings of the IEEE/CVF international conference on computer vision},
  pages={16259--16268},
  year={2021}
}

@misc{mmdet3d2020,
    title={{MMDetection3D: OpenMMLab} next-generation platform for general {3D} object detection},
    author={MMDetection3D Contributors},
    howpublished = {\url{https://github.com/open-mmlab/mmdetection3d}},
    year={2020}
}

@inproceedings{liu2021swin,
  title={Swin transformer: Hierarchical vision transformer using shifted windows},
  author={Liu, Ze and Lin, Yutong and Cao, Yue and Hu, Han and Wei, Yixuan and Zhang, Zheng and Lin, Stephen and Guo, Baining},
  booktitle={Proceedings of the IEEE/CVF international conference on computer vision},
  pages={10012--10022},
  year={2021}
}

@article{ma2024efficient,
  title={Efficient Modulation for Vision Networks},
  author={Ma, Xu and Dai, Xiyang and Yang, Jianwei and Xiao, Bin and Chen, Yinpeng and Fu, Yun and Yuan, Lu},
  journal={arXiv preprint arXiv:2403.19963},
  year={2024}
}

@inproceedings{mao2021voxel,
  title={Voxel transformer for 3d object detection},
  author={Mao, Jiageng and Xue, Yujing and Niu, Minzhe and Bai, Haoyue and Feng, Jiashi and Liang, Xiaodan and Xu, Hang and Xu, Chunjing},
  booktitle={Proceedings of the IEEE/CVF international conference on computer vision},
  pages={3164--3173},
  year={2021}
}

@article{wang2023sat,
  title={SAT-GCN: Self-attention graph convolutional network-based 3D object detection for autonomous driving},
  author={Wang, Li and Song, Ziying and Zhang, Xinyu and Wang, Chenfei and Zhang, Guoxin and Zhu, Lei and Li, Jun and Liu, Huaping},
  journal={Knowledge-Based Systems},
  volume={259},
  pages={110080},
  year={2023},
  publisher={Elsevier}
}

@article{song2023vp,
  title={VP-Net: Voxels as points for 3-D object detection},
  author={Song, Ziying and Wei, Haiyue and Jia, Caiyan and Xia, Yongchao and Li, Xiaokun and Zhang, Chao},
  journal={IEEE Transactions on Geoscience and Remote Sensing},
  volume={61},
  pages={1--12},
  year={2023},
  publisher={IEEE}
}

@article{liu2024sparsedet,
  title={Sparsedet: a simple and effective framework for fully sparse lidar-based 3D object detection},
  author={Liu, Lin and Song, Ziying and Xia, Qiming and Jia, Feiyang and Jia, Caiyan and Yang, Lei and Gong, Yan and Pan, Hongyu},
  journal={IEEE Transactions on Geoscience and Remote Sensing},
  year={2024},
  publisher={IEEE}
}
}

\clearpage
\clearpage
\maketitlesupplementary


\section{Implementation details}
\subsection{Network architecture}

For Argoverse2 and Waymo Open, we adopted the sparse version of CenterPoint~\cite{yin2021center} as the detection head, as used in~\cite{zhang2024safdnet}. 
For nuScenes, we employed TransFusion~\cite{bai2022transfusion} as the detection head, in line with~\cite{zhang2024hednet,wang2023dsvt}. 

\noindent
\textbf{Argoverse2.}
We begin by validating the effectiveness of our detector in the long-range detection of Argoverse2. 
The voxel sizes are set to $(0.1,0.1,0.2)$ meters.
In stages 1 through 3, we employ $(0, 1, 1)$ SFM blocks, each followed by $(2, 2, 4)$ SRBs, respectively. 
In stage 4 and within the 2D backbone, we use $(4, 2)$ SFM blocks, each followed by $(2, 4)$ SRBs. 
For each SFM block, we set $L=4$ with kernel sizes of $(3, 3, 3, 3)$ and dilation rates of $(1, 3, 5, 7)$. 

\noindent
\textbf{Waymo Open.}
For the mid-range dataset, we set the voxel sizes to $(0.08, 0.08, 0.15)$ meters. 
In stages 1 through 3, we employ $(0, 1, 1)$ SFM blocks, each followed by $(2, 2, 4)$ SRBs, respectively. 
In stage 4 and within the 2D backbone, we use $(2, 2)$ SFM blocks, each followed by $(6, 6)$ SRBs. 
Since the Waymo scanning range is not as large as that of Argoverse2, we set $L=4$ with kernel sizes of $(3, 5, 3, 5)$ and dilation rates of $(1, 1, 3, 3)$ for each SFM block. 
This configuration results in an effective receptive field (ERF) of $2$ meters, compared to $3.3$ meters in Argoverse2.

\noindent
\textbf{nuScenes.}
For the short-range detection dataset, we set the voxel sizes to $(0.075, 0.075, 0.2)$ meters and used the same configuration of SFM blocks and SRBs as in Waymo Open. 
Since nuScenes focuses on short-range detection (up to $\pm 54$ meters), it was shown in~\cite{zhang2024safdnet} that a sparse detection head offers no advantage over a dense one, as used in~\cite{zhang2024hednet}. 
Consequently, we implemented the 2D backbone with dense convolutions instead of sparse ones.

\subsection{Training and inference schemes}
We implemented our approach using the OpenPCDet~\cite{openpcdet2020} framework for the Argoverse2 and Waymo Open datasets and the MMDetection3D~\cite{mmdet3d2020} framework for nuScenes, following the setup of state-of-the-art methods~\cite{wang2023dsvt,zhang2024hednet,zhang2024safdnet}.

\noindent
\textbf{Argoverse2.}
We follow the same training schemes as~\cite{yin2021center} to optimize the network using the Adam optimizer with a weight decay of $0.05$ and a one-cycle learning rate policy. 
The maximum learning rate is set to $5\times10^{-3}$, and training is performed with a batch size of $32$ for $24$ epochs on $8$ NVIDIA L40 GPUs. 
Following~\cite{zhang2024safdnet}, we apply ground-truth copy-paste data augmentation during training, disabling this augmentation in the final epoch as part of a fade strategy. 
For reporting frames per second (FPS) during inference, we use a single NVIDIA L40 GPU.

\noindent
\textbf{Waymo Open.}
As in Argoverse, we follow the same training schemes as~\cite{yin2021center} to optimize the network using the Adam optimizer with a weight decay of $0.05$, a one-cycle learning rate policy, and a maximum learning rate of $5\times10^{-3}$.
Training is performed with a batch size of $32$ for $24$ epochs on $8$ NVIDIA A100 GPUs.
Following~\cite{wang2023dsvt,zhang2024hednet,zhang2024safdnet}, we apply ground-truth copy-paste data augmentation during training, disabling this augmentation in the final epoch as part of a fade strategy. 
During inference, and as described in~\cite{wang2023dsvt,zhang2024hednet,zhang2024safdnet}, we use class-specific NMS with IoU thresholds of $0.7$, $0.6$, and $0.55$ for vehicles, pedestrians, and cyclists, respectively.

\noindent
\textbf{nuScenes.}
We follow the training scheme adopted in~\cite{bai2022transfusion}. 
The network is trained using the AdamW optimizer with a weight decay of $0.01$, a one-cycle learning rate policy, a maximum learning rate of $1\times10^{-3}$, and a batch size of $32$ for $20$ epochs on $8$ NVIDIA A100 GPUs. 
Similarly to~\cite{bai2022transfusion,wang2023dsvt} we apply the fade strategy in the final $5$ epochs.


\begin{table}[t]
    \footnotesize
    \setlength\tabcolsep{4pt} 
    \centering
    \begin{tabular}{c c c c | c c }
         \toprule
            \multirow{2}{*}{Stage2} & \multirow{2}{*}{Stage3} & \multirow{2}{*}{Stage4} & \multirow{2}{*}{2D backbone} & \multicolumn{2}{c}{mAP/mAPH} \\
            & & & & L1 & L2 \\
        \midrule
            & & & & 75.6/73.3 & 69.0/66.9 \\
            & & & \cmark & 76.4/74.2 & 70.1/67.9 \\
            & & \cmark & \cmark & \textbf{78.0/75.7} & \textbf{71.7/69.5} \\
            & \cmark & \cmark & \cmark & 77.5/75.2 & 71.2/69.0 \\
            \cmark & \cmark & \cmark & \cmark & 77.9/75.6 & 71.6/69.4 \\
         \bottomrule
    \end{tabular}
    \caption{
    \textbf{Comparison of SFM module placement.}
    We integrate a single SFM module into various stages of the 3D backbone and the sole stage of the 2D backbone in our detector, evaluating on a $20\%$ subset of the Waymo Open dataset.
    The results indicate that focusing on stage $4$ of the 3D backbone and the 2D backbone is generally sufficient for optimal performance.
    }
    \label{tab:sfm_ablation_wod_sup}
\end{table}

\section{Additional ablation studies}
To save computational resources, we conducted ablations on $20\%$ of the Waymo Open dataset~\cite{waymo}, training for $12$ epochs on $8$ GPUs with a total batch size of $32$ and reporting results on the full validation set. 
We used a tiny SFMNet with a single SFM module in the 3D backbone and another in the 2D backbone, followed by $2$ and $4$ SRBs, respectively. 
To study SFM placement, we positioned the module at different backbone stages, and \tabref{tab:sfm_ablation_wod_sup} shows that placing it in stage $4$ of the 3D backbone and in the 2D backbone is generally sufficient.

\end{document}